\documentclass[preprint,12pt]{elsarticle}
\usepackage{amssymb}

\usepackage{amsmath}

\usepackage{mathtools}
\usepackage{amsthm}

\usepackage{algorithm}
\usepackage{algorithmic}
\usepackage{comment}
\usepackage{xspace}
\usepackage{enumitem}
\usepackage{graphicx}
\usepackage{makecell}

\usepackage{tikz}
\usepackage{amsmath}
\usepackage{longtable}
\usepackage{adjustbox}
\usepackage{enumitem}

\usepackage{rotating}
\usepackage{comment}
\usepackage{graphicx}
\usepackage{array}
\usepackage{booktabs} 
\usepackage{multirow}
\usepackage{caption}

\usepackage{graphicx}
\usepackage{lscape}
\usepackage[T1]{fontenc}
\usepackage[utf8]{inputenc}
\usepackage{xcolor}
\usepackage{tcolorbox}
\usepackage{enumitem}

\usepackage{xcolor}            
\usepackage{tabularx}
\usepackage{pifont}

\usepackage{multicol}
\usepackage{enumitem}

\journal{Applied Soft Computing}

\begin{document}

\begin{frontmatter}

\title{\Large \bf FRAME : Comprehensive Risk Assessment Framework for Adversarial Machine Learning Threats}

\author[1]{Avishag Shapira\corref{cor1}}
\ead{shavish@post.bgu.ac.il}

\author[1]{Simon Shigol}
\ead{shigols@post.bgu.ac.il}

\author[1]{Asaf Shabtai}
\ead{shabtaia@bgu.ac.il}

\cortext[cor1]{Corresponding author}

\affiliation[1]{organization={Software and Information Systems Engineering, Ben-Gurion University of the Negev},
                city={Beer-Sheva},
                postcode={8410501},
                country={Israel}}


\begin{abstract}

The widespread adoption of machine learning (ML) systems increased attention to their security and emergence of adversarial machine learning (AML) techniques that exploit fundamental vulnerabilities in ML systems, creating an urgent need for comprehensive risk assessment for ML-based systems.
While traditional risk assessment frameworks evaluate conventional cybersecurity risks, they lack ability to address unique challenges posed by AML threats.
Existing AML threat evaluation approaches focus primarily on technical attack robustness, overlooking crucial real-world factors like deployment environments, system dependencies, and attack feasibility. Attempts at comprehensive AML risk assessment have been limited to domain-specific solutions, preventing application across diverse systems.
Addressing these limitations, we present FRAME, the first comprehensive and automated framework for assessing AML risks across diverse ML-based systems.
FRAME includes a novel risk assessment method that quantifies AML risks by systematically evaluating three key dimensions: target system's deployment environment, characteristics of diverse AML techniques, and empirical insights from prior research.  FRAME incorporates a feasibility scoring mechanism and LLM-based customization for system-specific assessments.
Additionally, we developed a comprehensive structured dataset of AML attacks enabling context-aware risk assessment.
From an engineering application perspective, FRAME delivers actionable results designed for direct use by system owners with only technical knowledge of their systems, without expertise in AML. We validated it across six diverse real-world applications, including e-commerce feedback scoring and email security system.
Our evaluation demonstrated exceptional accuracy and strong alignment with analysis by AML experts.
FRAME enables organizations to prioritize AML risks, supporting secure AI deployment in real-world environments.
\end{abstract}

\begin{keyword}
Machine Learning \sep Risk Assessment \sep Adversarial Learning
\end{keyword}

\end{frontmatter}

\section{\label{sec:intro}Introduction}
The increased deployment of machine learning (ML) systems in critical domains has made them an attractive target for attackers. 
Over the past decade, studies have consistently demonstrated that ML models are inherently vulnerable due to logical flaws in their underlying algorithms, which adversaries can exploit~\cite{chakraborty2021survey,vassilev2024aml}. 
Adversarial machine learning (AML) refers to the broader field focused on understanding and addressing threats to ML systems. 
AML attacks include techniques that exploit vulnerabilities in the learning process and the decision-making capabilities of ML models~\cite{carlini2017adversarial,szegedy2013intriguing,huang2011adversarial}. 
These attacks encompass diverse vectors, such as adversarial examples, data poisoning, and model extraction, posing significant risks to the reliability, security, and trustworthiness of ML systems and the organizations that rely on them~\cite{vassilev2024aml}.
Existing cybersecurity risk assessment frameworks~\cite{nist7946,ganin2020multicriteria}, while robust for conventional security threats, are not designed to address the unique challenges posed by AML threats.

The distinct nature of AML vulnerabilities, which exploit ML systems' learning and decision-making processes, necessitates specialized methodologies tailored to these threats.
Recent studies on AML threat evaluation have mainly focused on developing frameworks and libraries for assessing attacks' robustness and developing technical defenses~\cite{rauber2020foolbox,nicolae2018adversarial,guo2023comprehensive}. 
Tools like the Adversarial Robustness Toolbox (ART) ~\cite{nicolae2018adversarial} provide valuable resources for benchmarking adversarial attacks and measuring model resilience. However, they are primarily focused on isolated technical evaluations, overlooking deployment-specific factors like system dependencies, operational constraints, and attack feasibility for concrete threat actors in real-world scenarios.

Similarly, guidelines like those published by NIST~\cite{vassilev2024aml}, standardize terminology and highlight key threats but leave actionable risk prioritization to system owners.
Domain-specific AML risk assessments, such the assessment of the Open Radio Access Network (O-RAN) performed by Habler et al.~\cite{habler2022adversarial}, provide tailored insights for specific applications but lack the generalizability required for comprehensive risk evaluation across diverse ML systems.

The main goal of this research is to develop a practical and automated risk assessment framework tailored to ML-based systems. 
Unlike existing approaches, our framework, FRAME, evaluates and quantifies AML threats by considering the ML-based system’s deployment characteristics, the attributes of AML attack techniques, and insights from prior AML research.
With the exception of an initial questionnaire, all steps in FRAME are  automated.
In addition, FRAME is intended to support technically knowledgeable system owners, eliminating the need for expertise in adversarial machine learning.
The framework outputs a prioritized list of risks tailored for the evaluated ML-based system, delivering practical and actionable results that enable system owners to effectively focus their mitigation efforts.
Six diverse real-world use cases were used to validate FRAME's effectiveness in a comprehensive evaluation that highlighted its adaptability and practical value.
As part of this research, we also created a exhaustive information source (dataset) containing structured information on adversarial attacks, that not only enhances the ability to perform thorough and context-aware risk assessments but also provides researchers with structured data to advance the study of AML threats and defenses. 

\vspace{0.2cm}
\noindent The contributions of our work can be summarized as follows:
\begin{itemize}[topsep=0pt,noitemsep,leftmargin=*]
    \item \textbf{First comprehensive risk assessment framework:} 
    To the best of our knowledge, FRAME is the first practical and automated framework designed to assess the risks posed by AML threats across diverse ML systems. Unlike prior works that focused solely on evaluating attack robustness or technical defenses, our framework provides a versatile and scalable risk assessment solution. 
    \item \textbf{Detailed feasibility assessment and risk evaluation:} We introduce a novel methodology for assessing and quantifying the factors required to execute various AML threats. 
    This includes a detailed evaluation of the threat actor's skills and resources and the potential impact on the target system. 
    Our methodology enables the identification and prioritization of high-risk threats and supports effective countermeasure planning.
    \item \textbf{Empirical evaluation across domains:} 
    FRAME was thoroughly evaluated across six diverse real-world applications, demonstrating its effectiveness in identifying and prioritizing adversarial risks, with validation from both AML experts and system owners.
    \item \textbf{Comprehensive dataset for adversarial threats:} We provide a comprehensive dataset that includes structured information on a wide range of adversarial attacks. 
    This dataset will serve as a valuable resource for performing detailed, context-aware risk assessments and enhance the ability to mitigate emerging threats in the AML landscape.

\end{itemize}

\section{\label{sec:background}Adversarial Machine Learning}
Adversarial machine learning (AML) refers to the study of the attacks and defenses on ML algorithms~\cite{vassilev2024aml}.
With the increasing deployment and popularity of ML in critical domains, AML has become an essential area of research, addressing vulnerabilities that arise throughout the ML lifecycle.
AML attacks can be categorized based on their objectives, stages, specificity, and attacker knowledge and by the execution mode.
Detailed description of AML attacks categorization is presented in~\ref{sec:background}.

\section{\label{sec:related_work}Related Work}

AML has become a critical area of research, driven by the increasing deployment of ML systems in multiple domains~\cite{vassilev2024aml}. 
While significant progress has been made in evaluating AML threats and developing mitigation strategies~\cite{chakraborty2021survey,vassilev2024aml}, the proposed approaches primarily focus on practical attack robustness and technical defenses. 
In addition, traditional cybersecurity risk assessment frameworks~\cite{ganin2020multicriteria,ekstedt2023yet} lack the capacity to address the unique vulnerabilities introduced by ML systems.
In this section, we review three key areas: traditional risk assessment frameworks in cybersecurity, existing AML risk assessment methodologies, and domain-specific risk assessment for adversarial threats.

\subsection{Traditional Cybersecurity Risk Assessment Frameworks}
Numerous methodologies have been developed to assess cybersecurity risks in IT systems, focusing on structured evaluations of system vulnerabilities and providing decision-making tools for system owners~\cite{ahmed2022mitre,aksu2017cvss,nist7946,spring2021cvss,manocha2021security,10.1093/cybsec/tyaa005,calder2018nist,ganin2020multicriteria,ekstedt2023yet}. 
For instance, Aksu et al.~\cite{aksu2017cvss} introduced a CVSS-based method that leverages attack graphs and predefined metrics to quantify system vulnerabilities, providing a systematic approach to identifying weaknesses. 
Similarly, the NIST IR 7946 report~\cite{nist7946} provided practical guidance on implementing CVSS v2.0 for structured scoring.
Ganin et al.~\cite{ganin2020multicriteria,ekstedt2023yet} introduced a multicriteria decision-making framework for assessing cybersecurity risks, integrating both quantitative metrics and qualitative evaluations. 
The MITRE ATT\&CK framework~\cite{ahmed2022mitre} extends traditional methodologies into a hybrid risk assessment approach that models adversarial tactics, techniques, and procedures (TTPs). 
While these approaches consider environmental characteristics and dependencies within traditional IT systems and effectively address conventional cybersecurity risks, they have a shared limitation when applied to ML-based systems, since they do not consider the unique vulnerabilities introduced by these systems.

\subsection{AML Risk Assessment}
The growing attention to adversarial threats has led to the development of AML-specific frameworks that focus on the evaluation of AML attacks~\cite{vassilev2024adversarial,rosenberg2021adversarial,liu2022mldoctor,nicolae2018adversarial,xu2020elephant,wu2021psc}. 
Several resources, such as the NIST AML guidelines published by Vassilev et al.~\cite{vassilev2024adversarial} and the taxonomy presented by Rosenberg et al.~\cite{rosenberg2021adversarial}, provide structured categorizations of AML attacks and defenses. 
These tools standardize terminology and highlight lifecycle vulnerabilities but lack methodologies for dynamic risk quantification. 
For instance, the NIST AML guidelines focus on attack definitions and defensive strategies but leave decision-making to system owners, without addressing deployment-specific goals or sensitive areas.
Practical evaluation tools like the ART~\cite{nicolae2018adversarial}, FoolBox~\cite{rauber2020foolbox}, and the evaluation framework proposed by Guo et al.~\cite{guo2023comprehensive} provide general-purpose methodologies for benchmarking the actual robustness of ML models against a range of adversarial attacks. 
Other frameworks like ML-DOCTOR~\cite{liu2022mldoctor} and the PSC framework~\cite{wu2021psc} focus on evaluating specific AML attacks; ML-DOCTOR emphasizes privacy-related attacks, such as membership inference, and the PSC framework centers on benchmarking adversarial attacks, primarily untargeted ones. 

While these frameworks provide valuable insights for the assessment of  technical robustness, they do not consider system-specific goals, operational constraints, or practical feasibility. 
Our framework integrates system-specific characteristics, operational constraints, and feasibility scoring into AML risk assessment, while incorporating insights from prior work that evaluate the practical robustness of AML attacks. 
This enables actionable, context-aware prioritization of adversarial threats, bridging the gap between technical evaluation and real-world risk management.

\subsection{Domain-Specific Risk Assessment in ML}
Several studies have examined AML risks within specific domains, providing tailored insights but limited generalizability across diverse applications~\cite{habler2022adversarial,ibitoye2023network,spie2021msc2,kandasamy2020iot}. 
For instance, Habler et al.~\cite{habler2022adversarial} analyzed adversarial threats in the O-RAN and proposed countermeasures for domain-specific vulnerabilities in telecommunications. 
Similarly, Ibitoye et al.~\cite{ibitoye2023network} surveyed AML risks in network security, aligning attack likelihood and severity using a grid-based mapping approach. 
In the IoT domain, Kandasamy et al.~\cite{kandasamy2020iot} presented a risk assessment framework emphasizing quantitative models for ranking IoT-specific threats, while Caballero et al. presented SPIE~\cite{spie2021msc2}, a framework that addresses robustness challenges in multi-source command and control systems, focusing on metrics for real-time data fusion under adversarial conditions. 

\noindent Despite their contributions, these studies focus narrowly on their respective domains and  limited numbers of attacks, lacking applicability to broader ML-based systems.
Our research addresses these gaps by providing a versatile, domain-agnostic risk assessment framework. 

\vspace{0.2cm}
\noindent\textbf{}In summary, while existing frameworks prospside valuable methodologies for cybersecurity risk assessment and adversarial ML evaluation, they often lack integration with system-specific goals, operational constraints, and adaptive AML threats. 
Our framework addresses these gaps by systematically assessing adversarial threats through structured analysis of their feasibility, system-specific dependencies, and potential impact on critical functions and sensitive areas. 
Additionally, it leverages empirical results from existing studies to enhance robustness evaluations, delivering actionable context-aware insights tailored to the needs of diverse stakeholders and system owners.

\section{Method}
\subsection{Design Principles}
FRAME is designed to address key principles that ensure its applicability, comprehensiveness, and utility for system owners. 
The key principles are:
   \textbf{Comprehensive coverage:} Covering all AML attacks, ensuring consideration of various stages (training or serving), objectives (integrity, availability, or privacy), and execution modes (digital or physical).
   
   \textbf{Use case-specific adaptability:} Tailoring the risk assessment to align with the unique properties, characteristics, and operational context of each evaluated system. 
   This includes accounting for the specific capabilities and knowledge of relevant concrete threat actors, ensuring that the assessment performed is both accurate and realistic for the system being evaluated.
   
   \textbf{Accurate and reliable evaluation:} Ensuring precise assessment of each attack technique's actual success rate and taking into account real-world constraints to deliver realistic and reliable threat evaluations.
   
   \textbf{Ease-of-use:} 
   Enabling effective and engaging use by system owners and stakeholders who have limited or no expertise in the AML domain.
   
   \textbf{Practical and actionable results:} Delivering meaningful, context-specific insights that system owners can easily interpret and use to effectively prioritize countermeasures.

\subsection{FRAME Overview}
\begin{figure}[h]
    \centering
    \includegraphics[width=1\linewidth]{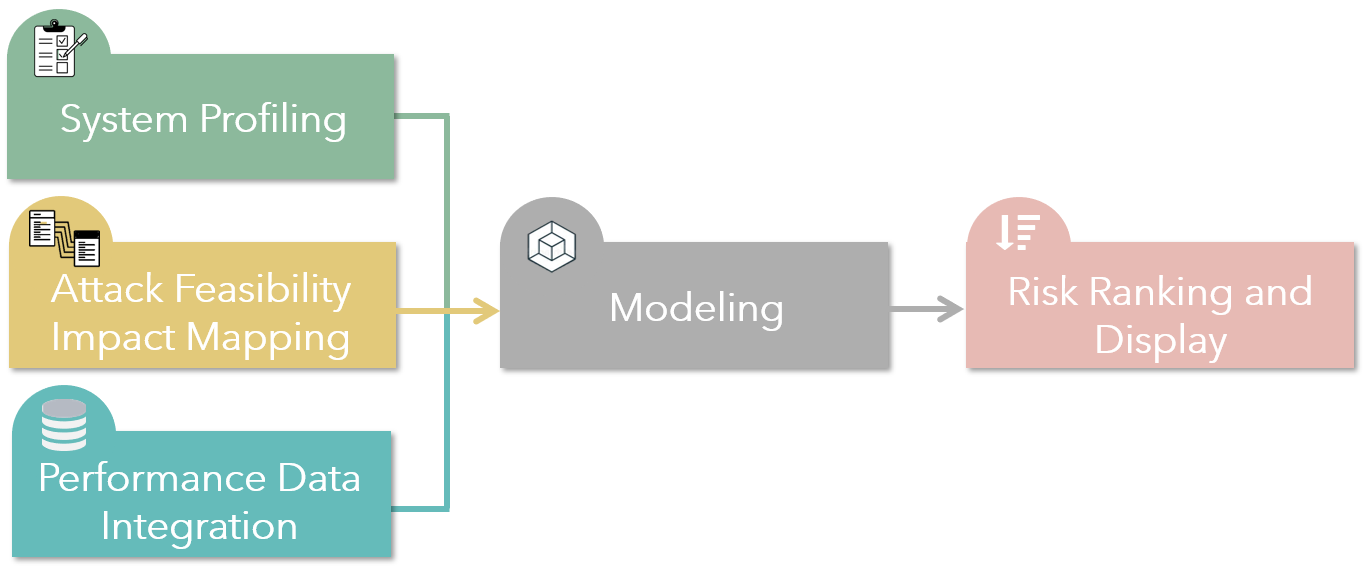}
    \caption{FRAME's main components.}
    \label{fig:flowOverviewl}
\end{figure}

Our framework, which is illustrated in Figure~\ref{fig:flowOverviewl}, consists of several components:

\begin{enumerate}[leftmargin=*]

    \item \textbf{System profiling: }
    This component utilizes a structured and questionnaire to collect detailed information about the evaluated ML-based system's characteristics.  
    The questionnaire is completed to profile the system, ensuring that the assessment is tailored to its unique attributes and contextualized for a selected threat actor. 
    This approach ensures \emph{use case-specific adaptability}.  
    Additionally, the questionnaire is designed to focus on system-specific information without requiring expertise in adversarial machine learning, while assuming technical familiarity with the evaluated system. 
    To further enhance usability, the questionnaire is automatically customized using an LLM based on the system description, making the questions easier to interpret and more relevant to the use case, thus supporting \emph{ease-of-use}.
    \item \textbf{Attack feasibility impact mapping: } 
    A pre-constructed mapping, developed with input from an AML expert, provides a detailed list of AML attacks and connects each attack to specific feasibility factors relevant for its execution and the security impacts it compromises.  
    By providing a comprehensive and detailed set of AML attacks for analysis, this mapping ensures \emph{comprehensive coverage}. 
    Once established, it functions as a reusable component across various system evaluations, offering tailored insights into the feasibility and impact of each attack, thereby contributing to \emph{accurate success evaluation}. 
    \item \textbf{Performance Data Integration (Dataset): }
    We created a structured and comprehensive dataset containing detailed records of AML attacks, compiled from academic literature and scientific publications. The dataset documents each attack's success rates under various conditions and captures its key characteristics.
    When evaluating an ML-based system, the framework identifies and retrieves relevant records from this dataset for each defined AML attack (listed in \emph{attack feasibility impact mapping} component), taking into account the specific characteristics of both the evaluated attack and the ML-based system.
    These records are used to empirically estimate the attack’s success rate, ensuring a data-driven and context-aware risk assessment.  
    By leveraging empirical evidence and context-specific data, this component supports \emph{accurate success evaluation} and enhances the framework's ability to deliver relevant and actionable insights.
    \item \textbf{Modeling}: 
    This component processes input from \emph{system profiling}, \emph{attack feasibility impact mapping}, and \emph{performance data integration} to compute a precise risk score for each of the AML attacks defined by \emph{attack feasibility impact mapping}, supporting \emph{accurate success evaluation}.   
    \item \textbf{Risk Ranking and Display:} 
    After computing the risk scores for all attacks during the \emph{modeling} process, this component ranks them in order of severity and presents the results in a structured, accessible format. 
    By providing practical insights, system owners and stakeholders can better understand the most critical threats to the evaluated system, aligning with the principle of \emph{practical and actionable results}. 

\end{enumerate}

While the system profiling questionnaire is completed manually by the user, all subsequent steps in FRAME, including attack matching, success rate retrieval, and risk scoring, are fully automated. 
To support usability, the manual stage is guided by an LLM-based customization process that adapts the questionnaire to the specific system under evaluation. This design ensures both adaptability and efficiency in the risk assessment process. 

\vspace{0.1cm}

\subsection{System Profiling}
FRAME is designed to be used by system owners who may not have expertise in adversarial machine learning, but who are familiar with the technical details of their system (e.g., model type, data flows, security posture). 

This component contains a structured questionnaire that profiles the ML system being evaluated. 
By collecting detailed system information, it enables a tailored risk assessment that considers both the system characteristics and safety. 
The questionnaire is completed each evaluated system and addresses a specific threat actor (e.g., an internal employee, customer, external attacker, or any other type of adversary). 
This ensures that the assessment reflects real-world conditions, as different threat actors might have varying capabilities, knowledge, and objectives.

This questionnaire consists of three parts: 
\noindent The \emph{System Characteristics} section gathers fundamental details about the ML-based system's use case, architecture, and data.
It provides a baseline understanding of the system's operational context.
Example questions include the Task Type (What is the ML model’s primary function? e.g., classification, regression, object detection), the Data Type (What type of data does the system process? e.g., image data, text data, time series), and the Model Architecture (What is the underlying model architecture? e.g., CNN, RNN, transformer).

\noindent The \emph{System Safety} section captures security-related properties that may influence the feasibility of adversarial attacks with respect to the chosen threat actor.
Example questions include the Task Type (What is the ML model’s primary function? e.g., classification, regression, object detection), the Data Type (What type of data does the system process? e.g., image data, text data, time series), and the Model Architecture (What is the underlying model architecture? e.g., CNN, RNN, transformer).
Example questions include the Trigger the Attack Digitally (How easy is it for the threat actor to manipulate the model’s inputs digitally at serving time? e.g., via an API) and the Model Online Evaluation (Does the system conduct ongoing evaluations? e.g., A/B testing).

\noindent The third section, \emph{Attack Impact}, captures the importance and potential consequences of breaching the key security principles of privacy, integrity and availability. 
This information helps prioritize risks by identifying which security principles are most critical for the evaluated system.
Example questions related to each principle include: How severe is the impact of an attack that causes the model to provide incorrect outputs for specific input? (Integrity), how critical is the exposure of training data or proprietary model components? (Privacy), and how significant is the impact of increased resource consumption during inference for specific inputs? (Availability).

Responses from the last two parts of the questionnaire are scaled between zero and one. For \emph{system security}, the scaling represents the degree to which conditions or properties relevant to attack feasibility are satisfied. For \emph{attack impact}, it indicates the severity of each impact in the specific use case. These scaled values are later mapped to feasibility factors and impacts.

\noindent The full questionnaire can be found in~\ref{Questionnaire}.

\vspace{0.1cm}
\textbf{Questionnaire Customization.}
To further enhance usability and improve ease of use for system owners, we introduce an automated customization process that tailors the system profiling questionnaire to the specific evaluated use case. This is achieved by leveraging a large language model (LLM), specifically ChatGPT, to adapt the original questionnaire using natural language system descriptions provided by the system owner (System Description). The customized questionnaire maintains the semantic structure of the original version while translating general questions into context-specific terms that are easier to interpret and answer,  leading to more accurate and consistent responses and reducing the likelihood of misinterpretation.  
Assuming the user has technical familiarity with the system, this process enables even those without adversarial ML expertise to provide reliable input.
An example of a customized questionnaire is included in~\ref{QuestionnaireCustomized}.

\subsection{Attack Feasibility Impact Mapping}
This component provides a structured approach for evaluating the feasibility and impact of AML attacks by linking attacks to conditions, capabilities, and system attributes (feasibility factors) required for its successful execution, as well as the security properties they compromise (impacts). 
Our mapping was created based on an extensive literature review, utilizing insights from comprehensive studies~\cite{vassilev2024aml,oprea2023adversarial,tian2022comprehensive,qin2023apbench,wang2022threats,gao2020backdoor} and was further refined with input from experts in the AML domain. 
This mapping consists of four elements: a list of attacks, list of feasibility factors, list of impacts, and attack feasibility impact mapping.

\noindent\textbf{List of Attacks. }
First, we define a set of AML attacks \( \mathcal{A} = \{a_1, a_2, \ldots, a_n\} \), where each \( a_i \) represents a predefined type of attack that could target ML systems. 
These attacks encompass a variety of techniques (e.g., data poisoning), threat models (e.g., white-box or black-box) and objectives compromised (privacy, integrity, or availability).
Examples include the White-Box Evasion/Misclassification Attack, that manipulates the input data and causes incorrect predictions while relying on full access to the model's parameters (e.g., adversarial examples designed for object detection models), and the Score-Based Membership Inference Attack, that enables to determine whether specific data points were part of the model's training dataset based on output probabilities.

\noindent The full list of attacks is provided in~\ref{AttackFeasibilityImpactMapping}. \\

\noindent\textbf{List of Feasibility Factors. }
Each attack depends on a distinct set of feasibility factors \( \mathcal{F} = \{f_1, f_2, \ldots, f_m\} \), where each \( f_j \) represents a  a specific condition, capability, or system attribute that influences an attacker's ability to successfully execute the attack.
Examples of feasibility factors include Manipulating the Training Data, which is the ability to alter or inject samples into the training dataset, and Model Feedback at Serving Time, which represents the level of feedback provided by the system during inference (e.g., only the final decision or the full probabilities of the output (score)).

\noindent The full list of feasibility factors can be found in~\ref{Capabilities List}.\\

\noindent\textbf{List of Impacts. }

\( \mathcal{I} = \{i_1, i_2, \ldots, i_p\} \) is the set of impacts, where each \( i_k \) represents the consequences of attacks, aligned with the core security objectives of privacy, integrity, and availability; each principle has multiple associated impacts.
These impacts help assess the harm an attack could cause to the system.
Each attack compromises one or more of the security impacts.
Example impacts include: Instance-specific incorrect output (Integrity), where the model is manipulated to produce incorrect results for specific attacker-chosen inputs, MIA (Privacy), where the attacker infers whether a particular data sample was part of the model’s training dataset, and Increased resource usage for all inputs (Availability), where the model is forced to consume excessive computational resources (e.g., energy, time) during inference.

\noindent The full list of security impacts can be found in~\ref{Impact List}.\\

\begin{figure}
    \centering
    \includegraphics[width=1\linewidth]{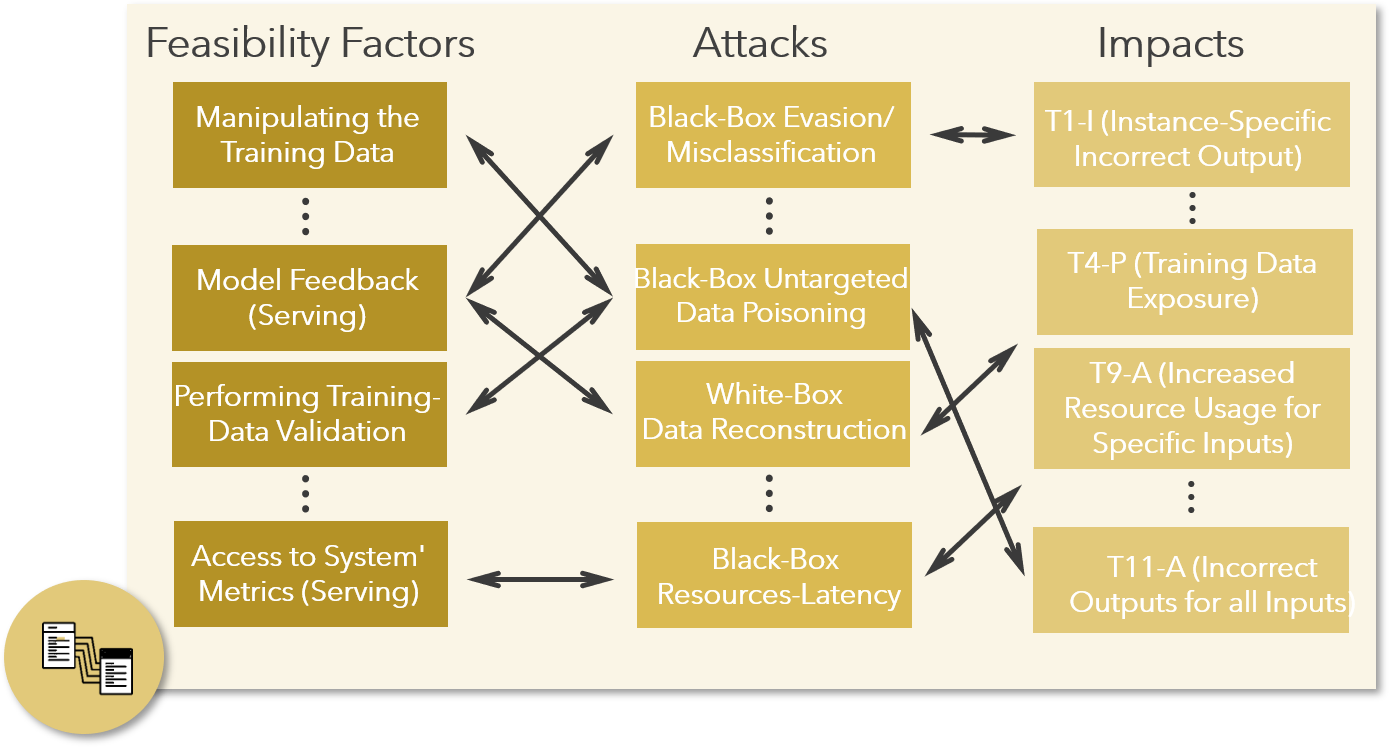}
    \caption{Attack feasibility impact mapping.}
    \label{fig:mapping}
\end{figure}

\noindent\textbf{Attack Feasibility Impact Mapping. } \label{AttCapMapping}
For each attack $ a\in A$, this mapping establishes the relationship between the attack, its required feasibility factors, and its potential security impacts.
An illustration of this mapping can be seen in Figure~\ref{fig:mapping}.
More formally, for each attack \( a \in A \), this mapping defines \( M_F(a) \), which is the subset of feasibility factors relevant successfully executing attack \( a \), and \( M_I(a) \) which s the subset of security impacts that are compromised by the attack \( a \):

\[
    M_F(a) = \{f \in F \mid f \text{ is relevant for the execution of } a\}.
    \]
 \[
    M_I(a) = \{i \in I \mid i \text{ is compromised by } a\}.
    \]

\noindent The full mapping between attacks, feasibility factors, and impacts is presented in~\ref{AttackFeasibilityImpactMapping}.

\subsection{\label{subsection:DB}Performance Data Integration (Dataset)}
This component provides the empirical foundation for robust risk assessment.
It leverages a comprehensive dataset collected from the academic literature on AML attacks to estimate how likely an attack would succeed against the evaluated system and the specific threat actor under consideration.

\noindent\textbf{Dataset Structure.} 
The dataset is composed of records, each representing a specific adversarial attack as described in the literature, covering a diverse range of attacks, domains, and use cases. 
Each entry includes detailed information about the use case, such as the domain of the attacked system (e.g., computer vision), the targeted models (e.g., YOLO), and the datasets used (e.g., MS-COCO). It also captures the characteristics of the attack, including the threat model (e.g., white-box or black-box) and the execution mode (e.g., physical or digital). Finally, each record documents the attack’s success rates, categorized as physical (e.g., adversarial patches in a physical environment) or digital (e.g., imperceptible perturbations applied to input data).

\subsection{Modeling}
\begin{figure*}
    \centering
    \includegraphics[width=0.94\linewidth]{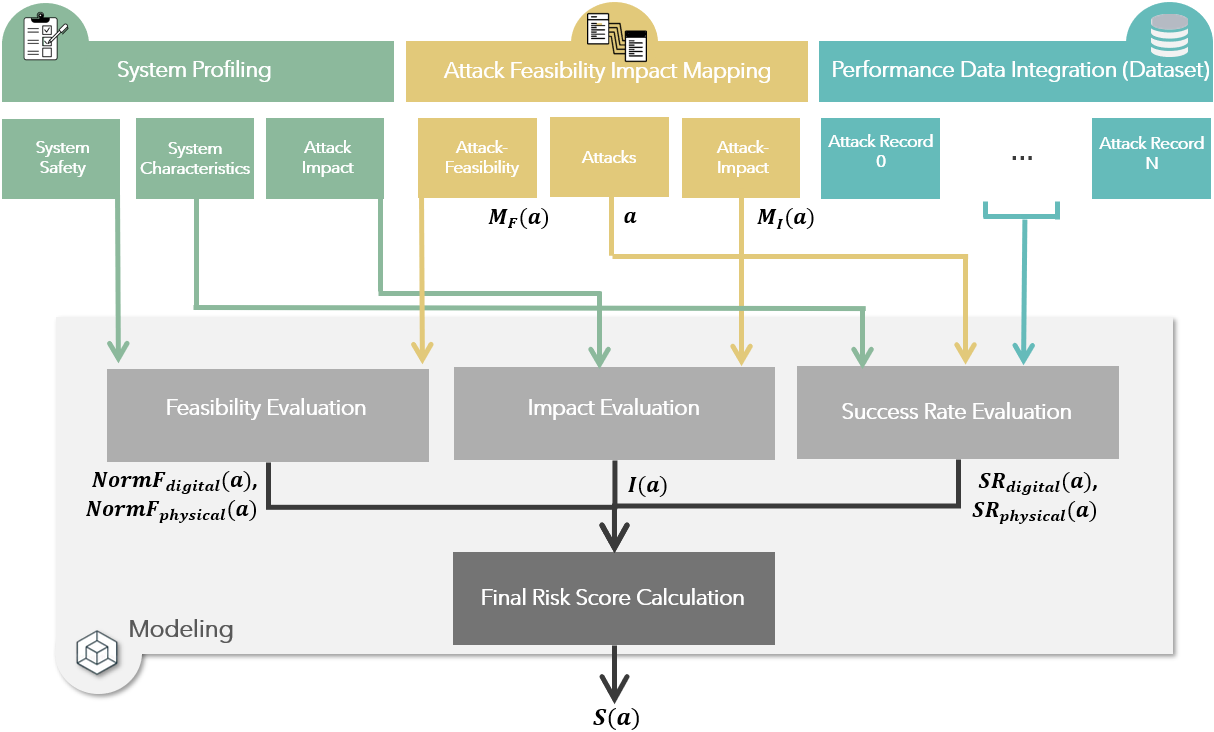}
    \caption{Modeling pipeline.}
    \label{fig:modeling}
\end{figure*}

The modeling component integrates inputs from the \emph{system profiling}, \emph{attack feasibility impact mapping} and \emph{performance data integration} to compute a comprehensive risk score $S(a)$ for each adversarial attack $s \in A$. $S(a)$ reflects both the likelihood of successfully executing the attack $a$ and its potential impacts on the evaluated system.
The process of calculating $S(a)$ for each attack $a$ can be seen in Figure~\ref{fig:modeling} and consists of the following steps: feasibility evaluation, impact evaluation, success rate evaluation, and final risk score calculation.

\subsubsection{Feasibility Evaluation}
The feasibility evaluation determines the extent to which an attack $a$ can be executed under the evaluated system's conditions. 
This is achieved by analyzing the relevant feasibility factors $M_F(a)$ as defined in the \emph{attack feasibility impact mapping}.
Each question in the \emph{system safety} part of the questionnaire is mapped to a specific feasibility factor $f$, with the response scaled to produce $Score(f)$, representing the feasibility of that factor.

The process is divided into the following stages:
 
\noindent\textbf{General Formula for Feasibility.}

The feasibility score $F_{generic}(a)$ for an attack 
$a$ is calculated by multiplying the scores of all relevant feasibility factors:
\begin{equation}
F_{{generic}}(a) = \prod_{f \in M_F(a)} {Score}(f)
\end{equation}

where $Score(f)$ represents the scaled response from the questionnaire for each factor $f$.

\noindent\textbf{Execution Mode-Specific Refinements}
Certain adversarial attacks can be executed through distinct execution modes, primarily categorized as \textbf{digital} or \textbf{physical}. These execution modes determine the specific mechanisms used to trigger the attack and interact with the target system.

The general feasibility score $F_{\text{generic}}(a)$ represents the combined feasibility of all general factors relevant to the attack $a$, however it excludes factors specific to the execution mode, such as those related to triggering the attack through digital or physical means.

To address these execution modes, we define two refined feasibility scores:
\begin{align}
{F}_{{digital}}(a) &= F_{{generic}}(a) \times {Score}({Factor}_{{digital}}) \label{eq:score_digital} \\
{F}_{{physical}}(a) &= F_{{generic}}(a) \times {Score}({Factor}_{{physical}}) \label{eq:score_physical}
\end{align}

where ${Factor}_{{digital}}$ and ${Factor}_{{physical}}$ are the factors representing the feasibility of triggering the attack through digital interaction (such as API access) and physical interaction (such as manipulating the system's environment) respectively.

\noindent\textbf{Scaling and Normalization}
In order to reduce the disproportionate influence of extremely high values in the final risk assessment, we apply a logarithmic scale on both ${F}_{{digital}}(a)$ and ${F}_{{physical}}(a)$.
To handle cases where the feasibility score is exactly zero, a small epsilon value ($\epsilon$) is added to ensure numerical stability during the logarithmic transformation.

After the logarithmic scaling, we apply min-max normalization to scale the scores back to a range of 0–1.
The final normalized feasibility score for each attack $a$ in both execution modes (digital and physical) is calculated as:
\begin{align}
NormF_{{em}}(a) &= \frac{\log(F_{{em}}(a)+\epsilon) - \min_{em}}{\max_{em} - \min_{em}} \label{eq:digital_context_normalized} 
\end{align}

where, $\min_{\text{em}}$ and $\max_{\text{em}}$ are the minimum and maximum values of $\log(F_{\text{em}}(a) + \epsilon)$ for all attacks $A$ in the respective execution mode (digital or physical).

A higher feasibility score results when an attack requires fewer factors and when the required factors are well-satisfied within the evaluated system.

\subsubsection{Impact Evaluation}
The impact evaluation determines the extent of damage an attack $a$ can cause to the evaluated system. 
This is achieved by analyzing the relevant impacts $M_I(a)$ as defined in the \emph{attack feasibility impact mapping}.
Each question in the \emph{attack impact} section of the questionnaire is mapped to a specific impact $i$, with the response scaled to produce $Score(i)$, representing the severity of that impact.

The relevant impacts for the attack \( a \), $M_I(a)$, are considered in the calculation of the total impact score for that attack:

\begin{equation}
I(a) = \prod_{i \in M_I(a)} \left( 1 - \text{Score}(i) \right)
\end{equation}

The total impact score $I(a)$ increases with both the number of impacts compromised and their severity to the system.

\subsubsection{Success Rate Evaluation}

This dataset contains documented adversarial attacks. Each entry (attack record) provides information about a specific AML attack described in previously published academic papers. The information includes the attack's characteristics, system properties (e.g., model architecture, domain), and observed success rates for applicable execution modes (digital and/or physical).
For each attack record, the success rate is divided into two values:

\begin{itemize}[topsep=0pt,noitemsep,leftmargin=*]
    \item \textbf{Digital Success Rate ($SR_{digital}(a)$):} The success rate when executing the attack digitally (e.g., adversarial perturbations applied to input data in a digital setting).
    \item \textbf{Physical Success Rate ($SR_{physical}(a)$):} The success rate when executing the attack in a physical environment (e.g., adversarial patches or physical-world sensor attacks), when applicable.
\end{itemize}

\textbf{Success Rate Estimation}
To estimate the success rate for each attack $a$, relevant records are retrieved from the dataset by matching the characteristics of $a$ (as defined in \emph{list of attacks}) and the evaluated system’s characteristics (as described in \emph{system characteristics}). 
This matching process, ensures that the most relevant dataset entries are selected to inform the success rate estimation.
Success rates are derived by performing two steps. 

First, the matching process searches for dataset records that align with the exact attack characteristics and system attributes. 
If exact matches exist, these success rates alone are used for estimation. When exact matches are unavailable, a downgrading process is applied. 
This involves easing the matching criteria step by step and grouping entries into batches based on their similarity to the evaluated conditions. 

Second, when applying the downgrading process, a weighted success rate formula is used to prioritize more relevant records from the dataset. 

For each attack \( a \), the success rate is calculated separately for the digital and physical execution modes, denoted as \( SR_{{digital}}(a) \) and \( SR_{{physical}}(a) \), respectively. 
Therefore, we apply this formula for both execution modes:

\begin{equation}
SR_{{em}}(a) = \frac{\sum_{i=1}^{n} B_{i,{em}} \cdot w_i}{\sum_{i=1}^{n} w_i}
\end{equation}

\( SR_{{em}}(a) \) is the success rate for the given execution mode (digital or physical) execution of attack \( a \).
 \( B_{i,{em}} \) is average success rate of all entries in the batch at downgrading level \( i \), specific to the given execution mode.
\( w_i \)is a predefined weight for downgrading level \( i \), prioritizing closer matches, and \( n \) is the total number of downgrading levels used in the calculation.

The weighted success rate calculation gives higher influence to closer matches, while still incorporating broader similarities with reduced weight.

FRAME’s downgrading strategy, described above, serves as a generalization mechanism, allowing the framework to estimate attack success rates even when exact dataset matches are unavailable. To further support coverage of novel systems and threats, the dataset used for success rate estimation is designed to be automatically expandable. As detailed later in Section~\ref{sec:Dataset Creation and Analysis}, the dataset creation component can be periodically triggered to scan new publications, ensuring continuous integration of emerging attack techniques, model architectures, and domains. This design enables FRAME to remain up to date with the evolving adversarial ML landscape.

\subsubsection{Final Risk Score Calculation}
The final risk score $S(a)$ integrates the feasibility, impact, and success rates for both digital and physical execution modes.

First, we calculate the likelihood of successfully applying $a$ for each execution mode (digital or 
physical):
\begin{equation} L_{{em}}(a) = NormF_{{em}}(a) \cdot SR_{{em}}(a)
\end{equation}

where $NormF_{em}(a)$ and $SR_{em}(A)$ are the feasibility score and success rate for the specific execution mode $em$.

Then, we combine the $L_{\text{em}}(a)$ of both execution modes to obtain the overall likelihood of successfully applying $a$:

\begin{equation} L_{overall}(a) = 1 - \prod_{\text{em} \in {\text{digital}, \text{physical}}} \left(1 - L_{\text{em}}(a)\right) \end{equation}

This formula ensures that the risk score reflects the contribution of both execution modes, increasing the final $S(a)$ if both modes are applicable.

Finally, to obtain the final score $S(a)$, we multiply over likelihood $L_{overall}(a)$ by the the total impact score of $a$, $I(a)$.
Additionally, for convenience and interpretability, $S(a)$ is scaled to a 0–10 range by multiplying it by 10. The maximum value of $S(a)$ is limited to 10 to avoid invalid score values:
\begin{equation} \label{finalEquation} S(a) = {Min}\left(L_{overall}(a)  \cdot I(a)\cdot 10, 10 \right) \end{equation}


\textbf{Zeroing $S(a)$ in Special Cases:}
When computing the final risk score, $S(a)$ becomes zero if any feasibility factor or the impact score is zero, due to the multiplicative calculation method. 
However, there are a few special cases where a feasibility factor is nonzero, yet the attack is inherently infeasible. 
For instance, in the \emph{model feedback - serving} factor, the feedback that the threat actor receives from the model at serving time is rated on a scale ranging from full access, score outputs (e.g., confidence scores), and decision outputs, to no feedback. If the system owner specifies "decision outputs" (a nonzero value), white-box and score-based attacks become infeasible, because these techniques rely on more detailed feedback (e.g., full access or score outputs).

To address such scenarios, we define a set of special cases with specific requirements for certain attacks. If these requirements are not met, such as particular values in critical feasibility factors, the $S(a)$ of the attack is explicitly set to zero. 
These special cases and their corresponding conditions can be found in~\ref{zeroingAppendix}.

\subsection{Risk Ranking and Display}
After modeling the risk scores $S(a)$ for all attacks $a \in A$, the attacks are sorted to produce a ranked list of risks. The highest-ranked attacks represent the most significant threats to the evaluated system and should be prioritized for mitigation and defense strategies.
An example of the framework's output can be seen in Figure~\ref{fig:top5}.

\noindent\textbf{Generating contextual scenarios for practical insights.}
To further assist system owners in understanding and mitigating these risks, we provide the top-five ranked attacks identified by the framework, along with the use case description and the associated threat actor, to a large language model (LLM) (specifically, ChatGPT-4o) to generate detailed descriptions of scenarios corresponding to each attack, specifically tailored to the discussed use case. 
This step transforms the framework output into a more practical and actionable result, enabling system owners to better grasp the real-world implications of the attacks and prioritize mitigation efforts effectively.

\section{Dataset Creation and Analysis}
\label{sec:Dataset Creation and Analysis}

\subsection{Dataset Creation}

To ensure the proposed risk assessment framework's robustness, we created a comprehensive dataset that provides empirical insights for evaluating AML attack success rates.

\begin{figure}[h]
    \centering
\includegraphics[width=1\columnwidth]{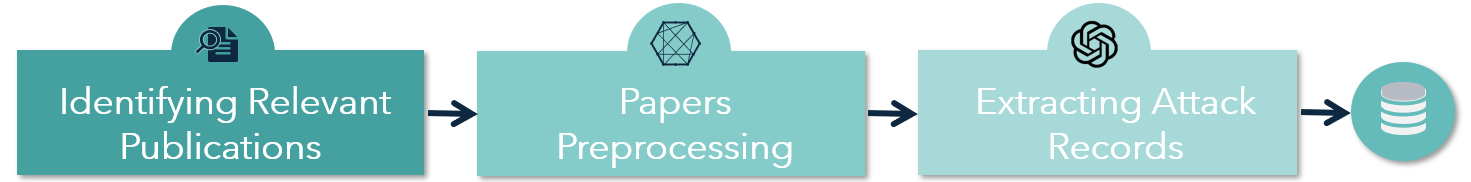}
    \caption{Dataset construction pipeline.}
    \label{fig:KnowledgeBaseFlow}
\end{figure}

\noindent\textbf{Dataset construction pipeline.}
Our dataset was created using a semi-automated pipeline designed to systematically collect and process data from academic publications. This approach ensures consistency, scalability, and the inclusion of data relevant to adversarial attacks.
The proposed pipeline consists of three steps, which are illustrated in Figure~\ref{fig:KnowledgeBaseFlow}.

\textit{1. Identifying relevant publications:}  
The pipeline begins by querying academic databases, using carefully designed search terms that combine attack types (e.g., evasion attack, poisoning attack), application domains (e.g., computer vision, cybersecurity), objectives, and other relevant criteria. 
The list of queries is provided in~\ref{Articles Query List}.  
To ensure that the dataset focuses on relevant and impactful work, we only include papers published from 2010 onward that appeared in high-impact conferences and journals (listed in the~\ref{ConAndJou}).  
Each publication is verified against existing entries to prevent duplicates, even when found through different queries.

\textit{2. Publication preprocessing:}  
Once relevant publications have been identified, the content is preprocessed and extracted. 
The documents are first processed to extract key sections, including text, tables, and images. 
The text is cleaned to remove irrelevant content (such as headers, footers, and page numbers), ensuring that only pertinent information remains. Tables are parsed into structured formats for easy analysis.
For images containing textual information,  the Tesseract Optical Character Recognition (OCR) engine~\cite{smith2007overview} is applied to extract any text from figures, graphs, and other visual elements.

\textit{3. Extracting attack records:}  
Following preprocessing, we extract key information from each publication.
To streamline this process, we utilized an LLM, specifically ChatGPT-4o.
We designed a structured prompt to guide the automated extraction of key information from the dentified publications, including: \textit{(i)} publication metadata (title, authors, year, publication name, and venue); \textit{(ii)} attack attributes (attack objectives, and characteristics, e.g., white-box/black-box, physical/digital execution modes); \textit{(iii)} evaluation context (attacked model's architectures, datasets, and evaluation environments); and \textit{(iv)} success rates (the success rates of the attack described in the paper in the digital and physical execution modes.
This prompt is specifically designed to extract a comprehensive list of all adversarial attacks presented in a paper, differentiating variations based on characteristics. 
For example, if a paper describes both white-box and black-box versions of an attack, which are evaluated across two domains (e.g., NLP and computer vision (CV)), the pipeline generates four distinct attack records. 
Each record corresponds to a unique combination of attack type and evaluation domain, ensuring coverage of all variations described in the publication.
The prompt directs the LLM to identify cases where publications are found to be irrelevant or do not describe valid adversarial attacks, and return that the publication is not relevant. In such instances, the paper is excluded from the dataset, ensuring that the dataset remains focused on high-quality, pertinent records. 

A validation prompt is applied to refine ambiguous outputs, standardize entries, and ensure compliance with predefined data quality criteria. 
This dual-prompt approach minimizes inaccuracies and improves data reliability.

\noindent\textbf{Dataset validation.}
To ensure the dataset's reliability, we manually validated 50 randomly selected records, assigning an accuracy score to each. The automated extraction process achieved an average accuracy score of 0.8 across these records, confirming its effectiveness in capturing accurate and comprehensive information.

\noindent\textbf{Contribution.}
This dataset is a cornerstone of our risk assessment framework, enabling precise estimation of attack success rates and their contextual relevance. 
In addition to its role in the framework, it will serve as a valuable resource to the broader research community, providing a centralized repository of adversarial attack insights. 
As new academic publications emerge, the dataset can be periodically expanded to include additional works by automatically performing the steps described above at regular intervals. This will ensure that the dataset remains up-to-date and continues to reflect the latest advancements in the field. 

\subsection{Key Insights}
To extract meaningful insights from our dataset, we conducted a comprehensive analysis examining attack trends, success rates, and domain-specific characteristics across the collected publications. 
Below, we present several key findings that emerged from this analysis, highlighting patterns that can enhance risk assessment methodologies and guide future research.

\begin{figure}[H]
    \centering
    \includegraphics[width=0.5\linewidth]{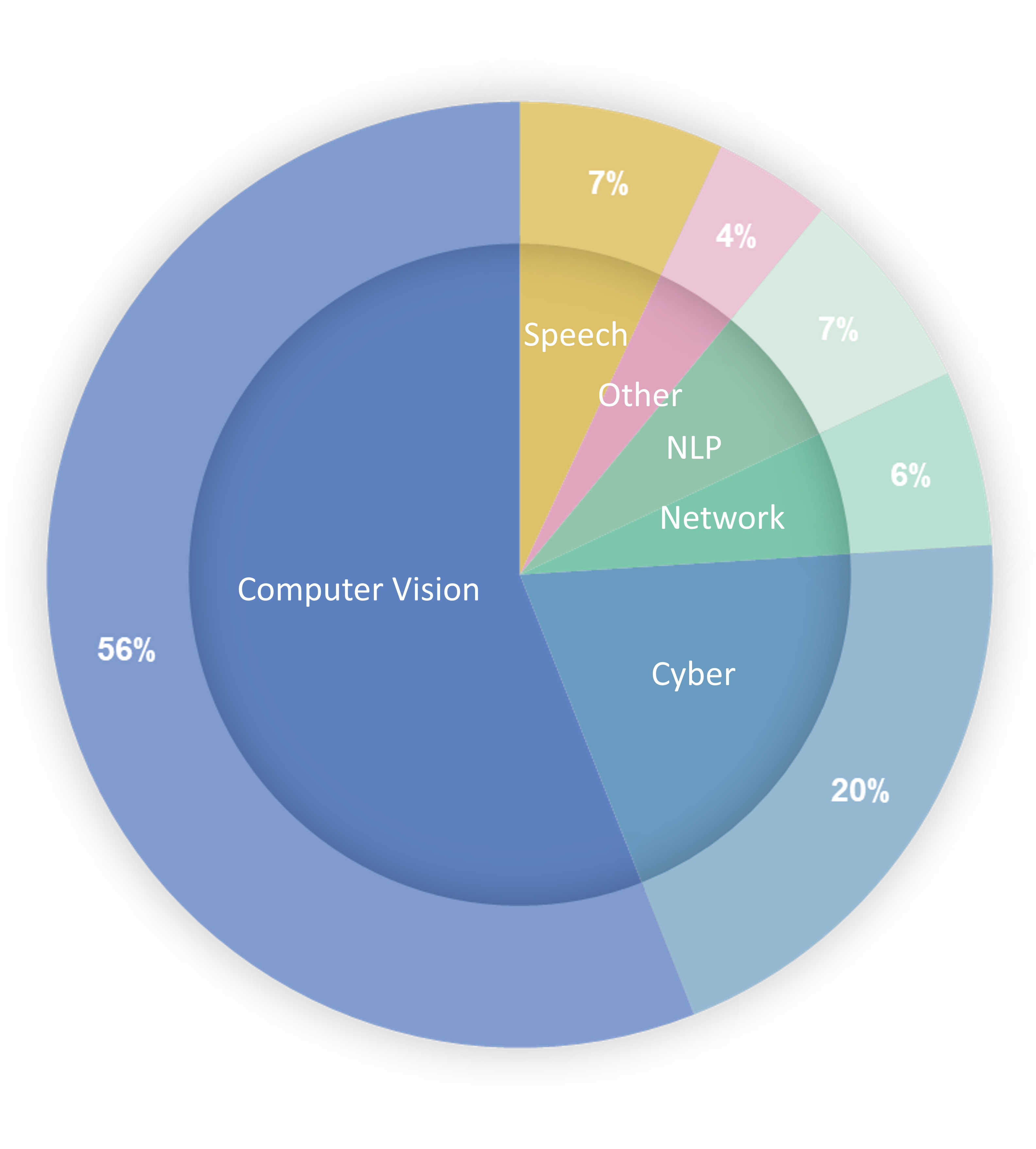}
    \caption{Distribution of AML research by domain.}
    \label{fig:AMLResearchDomains}
\end{figure}

\begin{figure}[H]
    \centering
\includegraphics[width=0.9\linewidth]{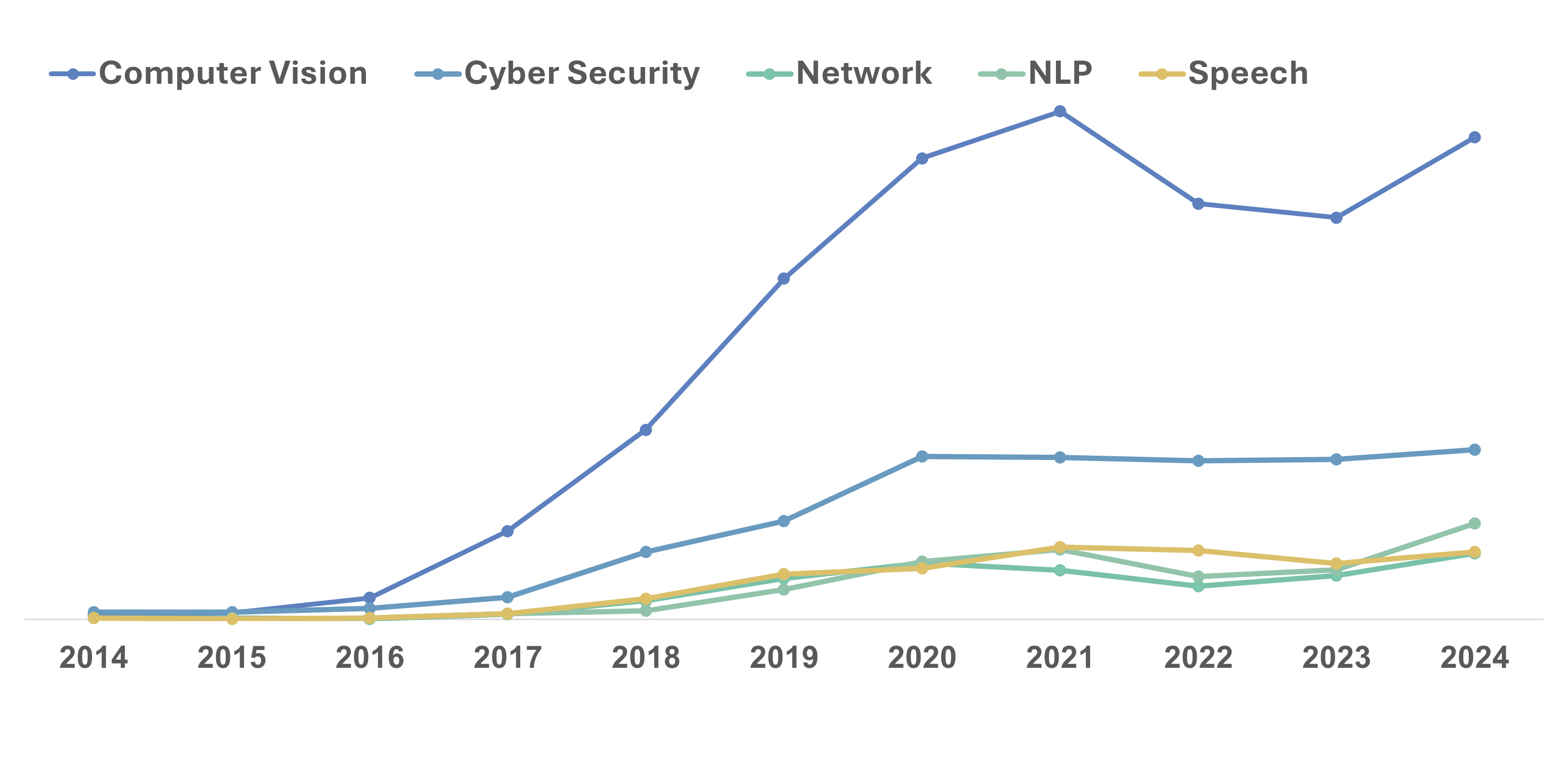}
    \caption{Articles over the years by domain.}
    \label{fig:ArticlesOverTheYearsbydomainAconJou}
\end{figure}

\noindent\textbf{Attack distribution by domain.} 
Several trends were identified in our analysis of the domain distribution in our dataset. As can be seen in Figure~\ref{fig:AMLResearchDomains}, we found that CV is the most studied domain, representing 56\% of academic publications in our dataset; this is followed by the cybersecurity domain which represents 21\% of the dataset's publications.
Figure~\ref{fig:ArticlesOverTheYearsbydomainAconJou} presents the publication trends in the various domains over the years, highlighting a significant rise in CV research starting in 2017 and peaking in 2021. During that period, the number of CV papers was \textit{3x} greater than those in the cyber domain and \textit{10x} greater than those in the NLP domain.

\begin{figure}[h]
    \centering
    \includegraphics[width=0.9\linewidth]{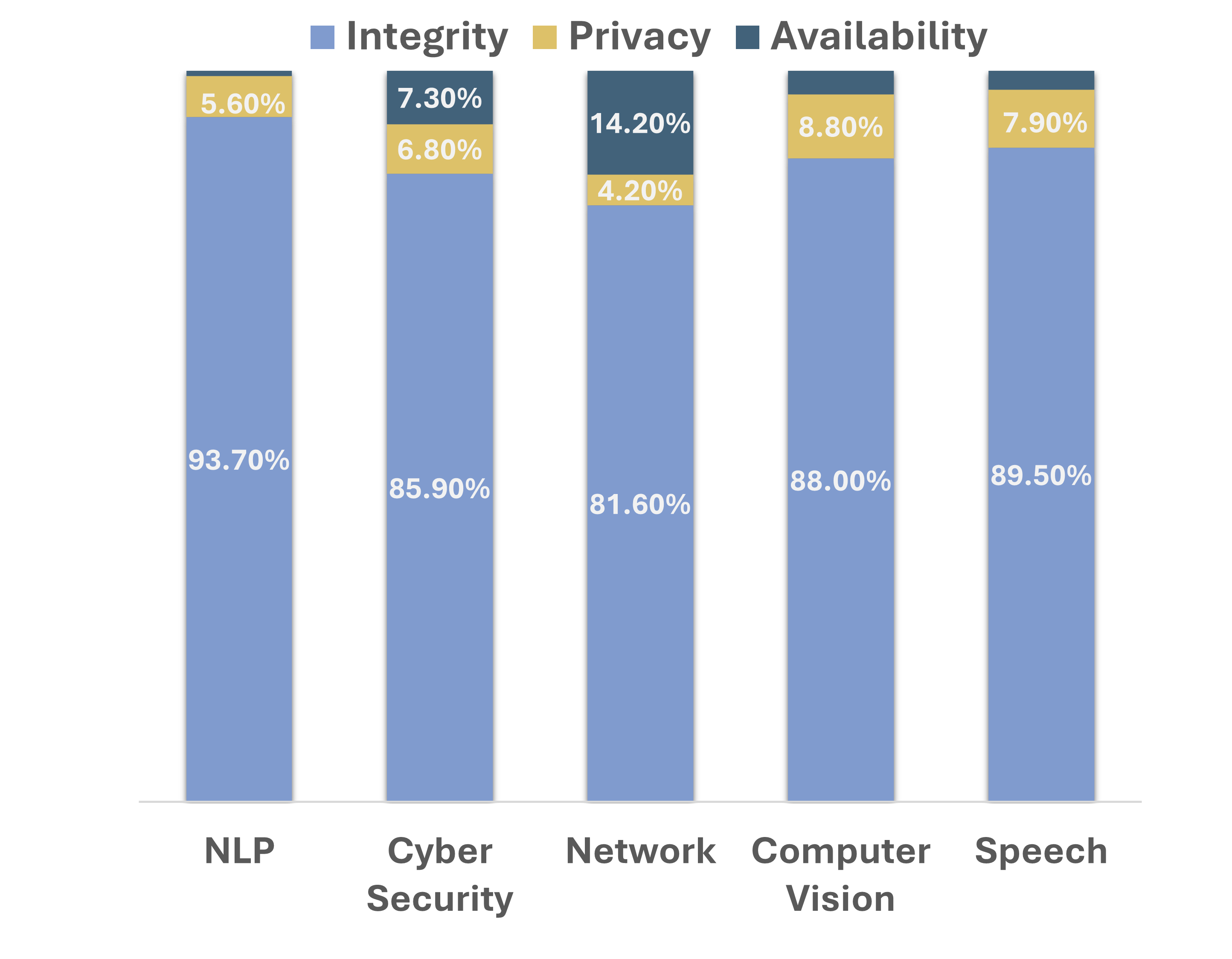}
    \caption{Distribution of attack objective by domain.}
    
    \label{fig:objective}
\end{figure}

\newpage
\noindent\textbf{Attack objective.} 
The distribution of attack objectives across application domains (Figure~\ref{fig:objective}) shows that \textit{integrity} attacks dominate across all domains, accounting for over 80\% of the dataset records. This dominance is particularly pronounced in the \textit{NLP} domain, where integrity attacks constitute 94\% of the records, with the remainder of the attacks primarily targeting privacy.
Although \textit{availability} attacks are less common (~4\% overall), they are notably more prevalent in the \textit{network} (14\%) and \textit{cyber} (7\%) domains.

\begin{figure}[h]
    \centering
    \includegraphics[width=0.9\linewidth]{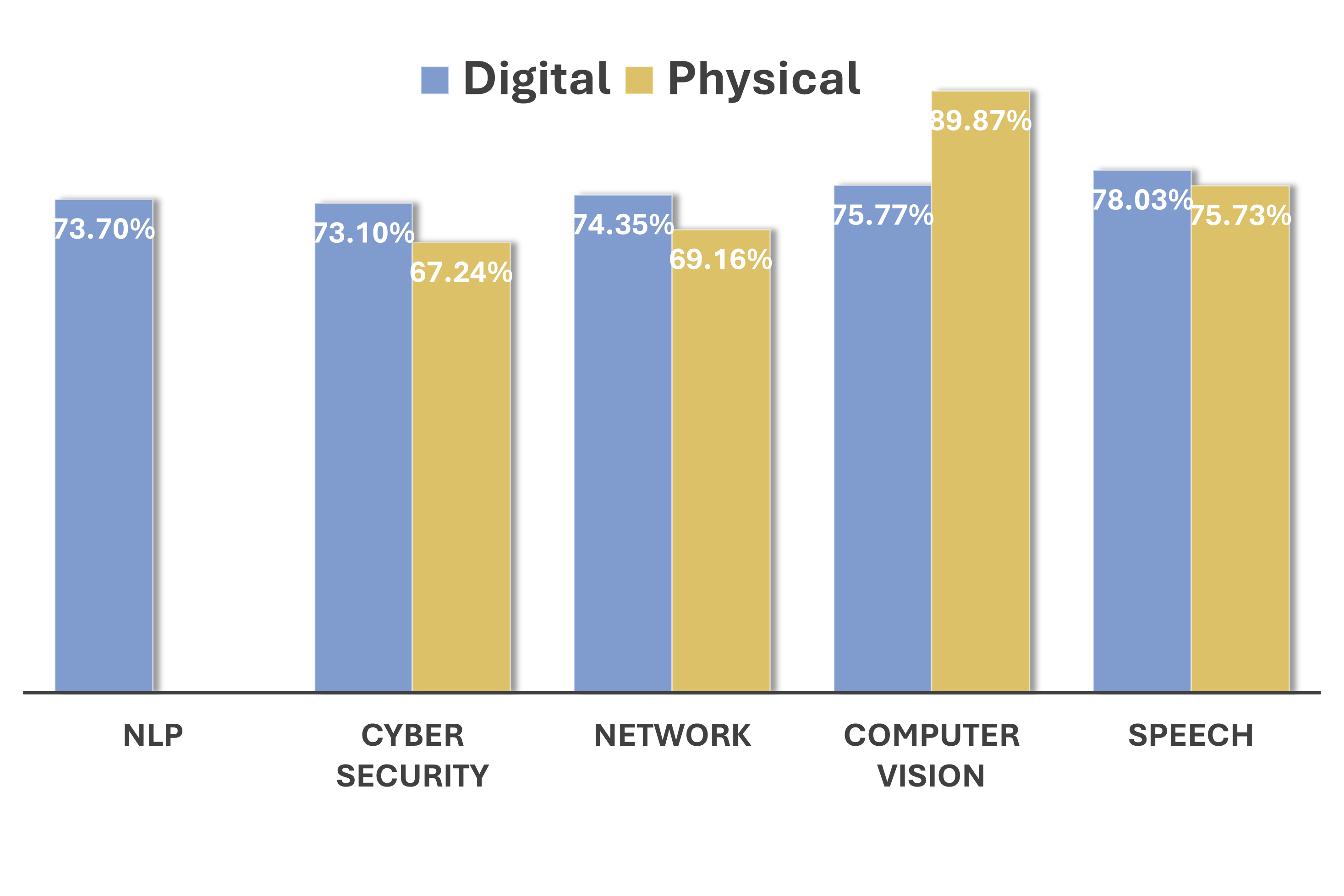}
    \caption{Success rate of physical and digital attacks.}
    \label{fig:SuccessRateByDomains}
\end{figure}

\begin{figure}[H]
    \centering
    \includegraphics[width=0.75\linewidth]{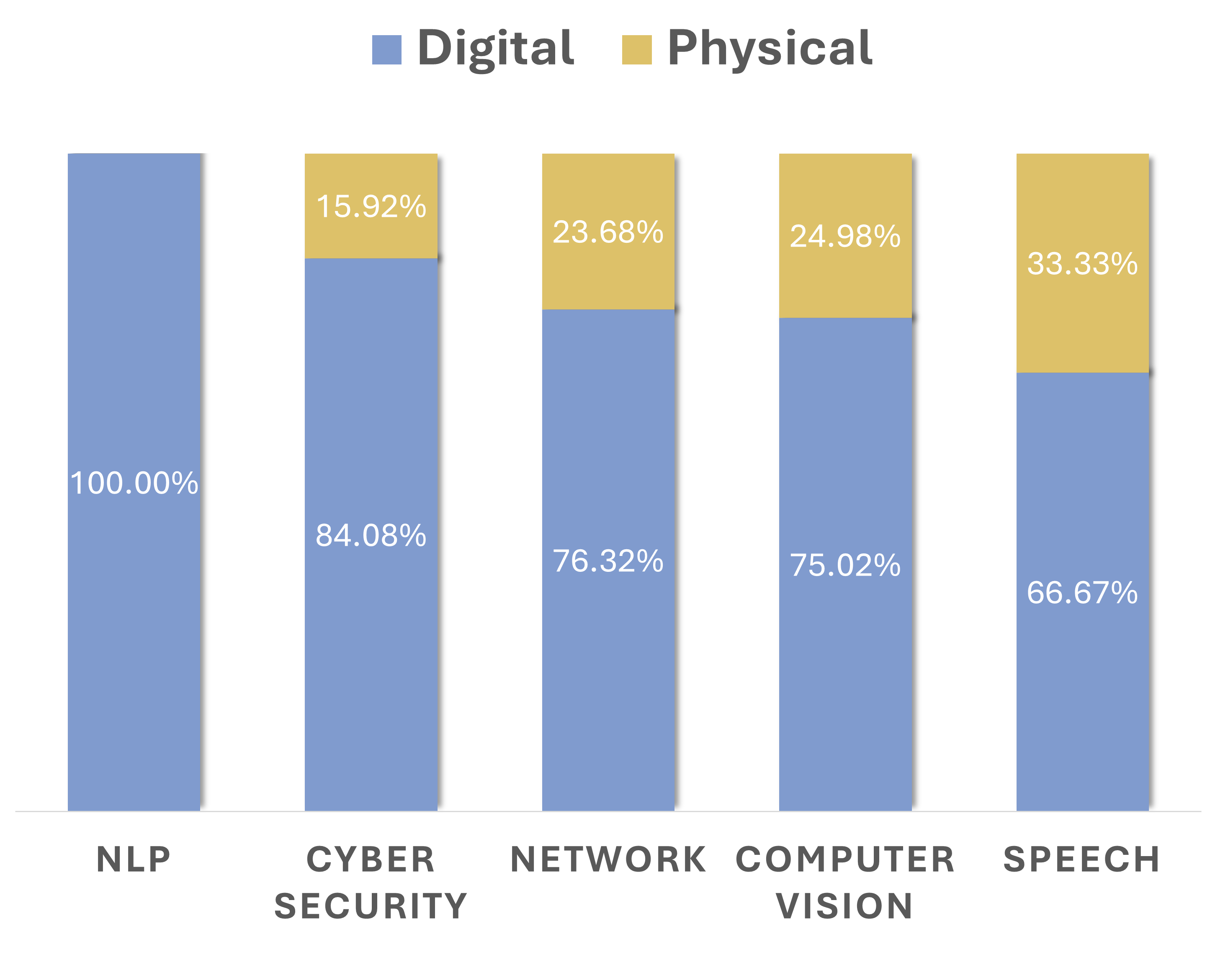}
    \caption{Ratio of physical and digital attacks in different domains.}
    \label{fig:DigPhyRatio}
\end{figure}

\noindent\textbf{Success rate by execution mode.}
As illustrated in Figure~\ref{fig:SuccessRateByDomains}, digital attacks consistently achieve higher success rates than physical attacks across all domains. As expected, no physical attacks were identified in the NLP domain in our dataset.

Our analysis of physical versus digital adversarial attacks across domains (Figure~\ref{fig:DigPhyRatio}) highlights the domain-specific nature of adversarial strategies. Physical attacks are particularly notable in the speech domain (14.5\%), potentially due to the ease with which acoustic signals can be manipulated in real-world settings (e.g., introducing carefully crafted noise or altering spoken commands to deceive speech recognition systems)~\cite{yakura2018robust}. Such attacks often exploit the physical properties of sound, making speech systems uniquely vulnerable to physical adversarial techniques.

In addition, our dataset indicates that physical variants of AML attacks were mainly observed in evasion/misclassification and poisoning attacks, which align with the nature of these attack. For example, evasion attacks often involve altering objects in a scene, such as modifying traffic signs to mislead autonomous vehicles or manipulating visual elements to confuse object detection systems~\cite{song2018physical,shapira2022attacking}. Similarly, audio-based evasion might include embedding adversarial noise into spoken commands to bypass authentication systems~\cite{yakura2018robust,chen2022push}. Poisoning attacks frequently involve tampering with sensor inputs or corrupting training data used by ML systems, such as introducing faulty sensor readings or maliciously crafted datasets~\cite{ino2024feasibility,ino2024data}.

\begin{figure}[h]
    \centering
    \includegraphics[width=0.75\linewidth]{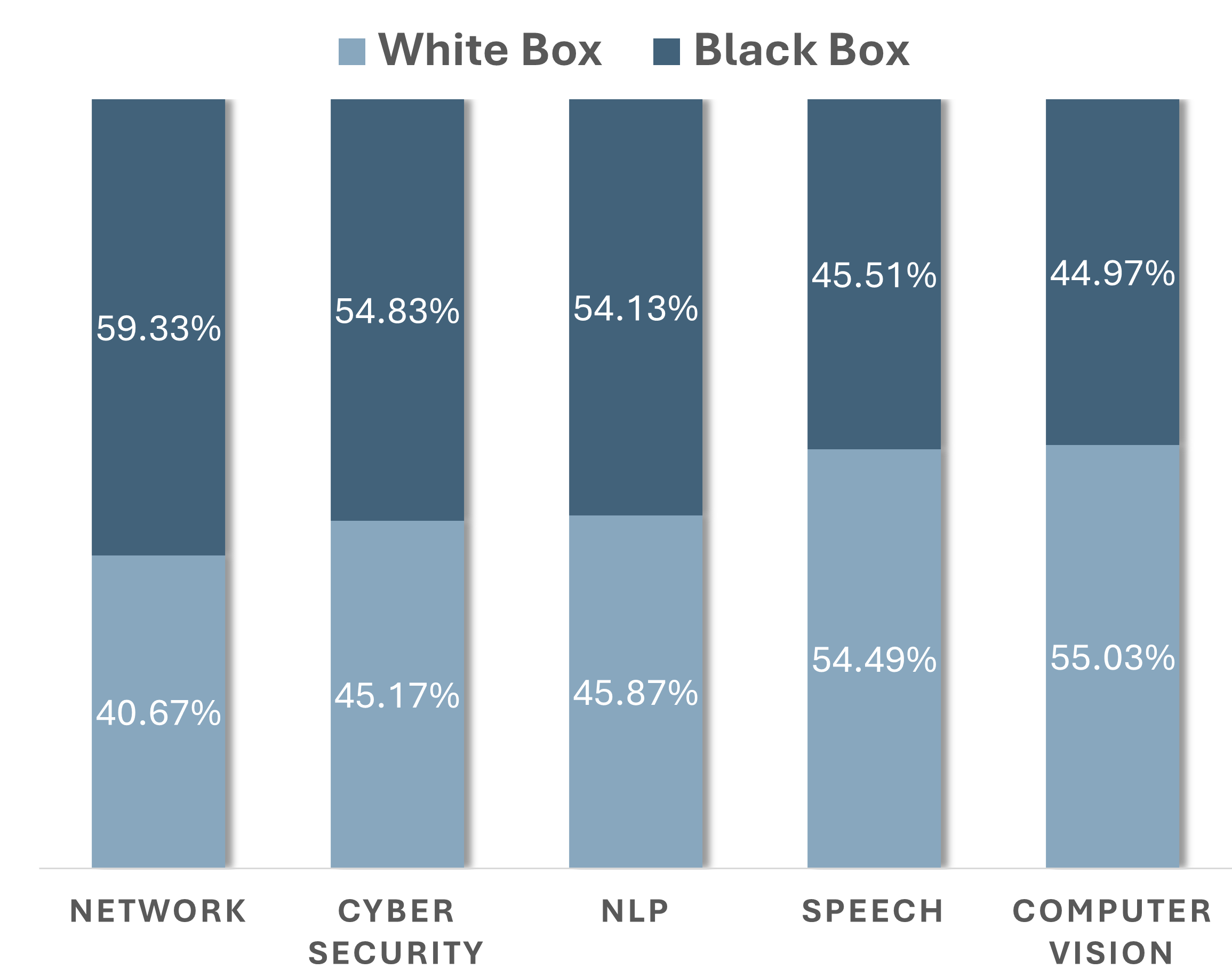}
    \caption{Attacker knowledge in different domains.}
    \label{fig:BBWB}
\end{figure}

\noindent\textbf{Black-box vs. white-box attacks.}
In the dataset there is a near-even split between white-box and black-box attacks, with white-box attacks constituting 52\% of the total records. 
However, as shown in Figure~\ref{fig:BBWB}, this distribution varies across domains.
In \textit{Computer Vision (CV)} and \textit{Speech}, white-box attacks dominate, accounting for approximately 55\% of the attacks. 
This trend can be attributed to the accessibility of model architectures and datasets in these domains, which enables adversaries to leverage comprehensive knowledge of the systems.
Conversely, in \textit{NLP} and \textit{Cyber}, black-box attacks are more prominent, comprising around 55\% of the recorded attacks. 
In the \textit{Network} domain, black-box attacks are particularly dominant, reaching up to 60\%. 
This trend may reflect the practical constraints adversaries face in accessing proprietary models and data in these domains, making black-box techniques more viable.

\section{\label{sec:eval}Evaluation}
To validate the effectiveness and practical utility of FRAME, we conducted a comprehensive evaluation, involving six real-world ML systems deployed by different organizations across diverse domains: (1) a user feedback Scoring Model, (2) a malware classification system for email security, (3) a traffic steering system in O-RAN, (4) an image quality ranking system, (5) a beam hopping system for satellite communication, and (6) a product relevance classification system.
The evaluation aimed to assess how well the framework identifies and prioritizes adversarial threats relevant to each system.

\subsection{Evaluation Methodology}

For each of the examined ML systems, the evaluation process consists of four stages.

First, system profiling was conducted by developers and researchers who work closely with each selected system. These participants completed the profiling questionnaire to provide the necessary input for FRAME. 

Next, the framework generated a ranked list of adversarial attacks for each system, based on their estimated likelihood and potential impact. 

In the third stage, the most critical attacks identified by FRAME were presented to the same developers and researchers, who were then asked to provide binary feedback (agree/disagree) on whether they considered these attacks to represent the most significant risks to their system. 

Finally, to further validate the framework’s output, a group of three AML experts collaboratively evaluated FRAME’s performance. After reviewing the system descriptions, the definition of the chosen threat actor, and the framework's outputs, the experts engaged in group discussions for each use case. Then they collectively completed a structured questionnaire assessing the framework's accuracy, relevance, and comprehensiveness.

\subsection{Evaluation Metrics for Each ML System}
For each ML system, we employed the following metrics:
\begin{itemize}
    \item \textbf{System Owners' Alignment:} This metric evaluates whether the top-five threats identified by FRAME align with the opinions of individuals closely involved with the ML system (e.g., developers, researchers). These individuals provided binary feedback (\textit{agree/disagree}) on whether FRAME correctly prioritized the most significant risks to their system.
    \item \textbf{Experts' Alignment}
    AML experts assessed FRAME using a structured questionnaire, focusing on four key aspects.
    \emph{Overall Framework Accuracy} measured how accurately FRAME’s ranking reflected real-world risk levels, rated on a scale from 1 to 10, with higher scores indicating stronger alignment with expert expectations.
    
    The \emph{Top Threat Relevance} evaluated the practical relevance of the top-five identified threats to the system, using a scale ranging from \emph{strongly agree} to \emph{strongly disagree}. This metric assesses the framework’s effectiveness in pinpointing realistic and significant risks.
        
    For \emph{Objective Coverage} experts rated FRAME's coverage of each core security objective (integrity, privacy, and availability) as \emph{yes}, \emph{partially}, or \emph{no}.
        
    Finally, \emph{Attack-Specific Accuracy and Relevance} involved detailed feedback for each of the top-five ranked attacks, where experts rated (on a scale of 1 to 10) both the \textbf{Accuracy} of the assigned score in representing the real-world severity of the attack, and its \textbf{Relevance}  in the context of the evaluated system and Whether the attack is practically meaningful and represents a significant threat in the context of the evaluated system.  
        
\end{itemize}

\subsection{Use Case Analysis}
In this section, we provide an in-depth analysis of a selected use case to demonstrate the effectiveness of FRAME. 
This analysis includes a description of the system and threat actor, the results generated by the framework (including detailed descriptions of the top-five attacks tailored by the LLM for the use case) and feedback received from system owners and AML experts. 
The other five use cases analyses are provided in~\ref{useCasesAppendix}.

\subsubsection*{Selected Use Case: Feedback Scoring Model}
\textbf{System Description:}  
A feedback scoring model designed for an e-commerce platform.
The model analyzes and scores the quality of buyer feedback reviews for the sellers' products sold on the platform.
Product reviews are displayed for potential buyers, according to their assigned score (from the the highest score to the lowest score), influencing purchasing decisions.
The scoring system considers various factors to determine a review's score, including its content quality, usefulness, and perceived relevance.

\noindent\textbf{Threat Actor Description:}  
The chosen threat actor is an attacker who has the ability to submit product reviews by posing as a legitimate buyer on the platform.

\begin{figure*}[h]
    \centering
\includegraphics[width=\linewidth]{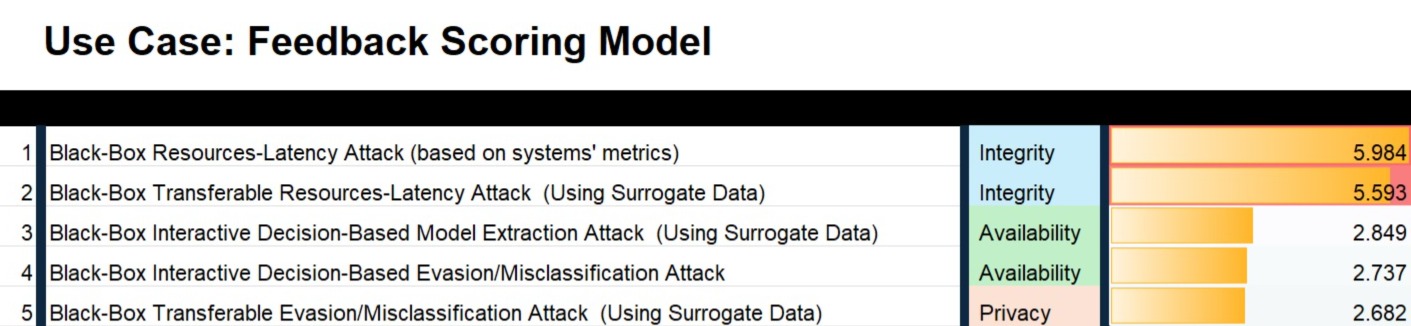}
    \caption{The top-5 attacks identified by our framework for the Feedback Scoring Model.}
    \label{fig:top5}
\end{figure*}

\noindent\textbf{Top Identified Attacks and Analysis:}
The top-five attacks are listed in Figure~\ref{fig:top5}.
\begin{enumerate}[topsep=0pt,noitemsep,leftmargin=*]
    \item \textbf{Black-Box Interactive Decision-Based Evasion/Misclassification Attack} (\textbf{Score:} 5.984) 
    \textbf{Objective:} Integrity \\
    \textbf{Explanation:} The attacker repeatedly submits variations of reviews and observes which ones receive higher rankings (based on where the reviews are displayed in the review list). This iterative process allows them to refine their submissions until they achieve their desired outcome: displaying favorable product feedback reviews or harming a seller's product feedback reviews. By exploiting the model's ranking mechanism, attackers manipulate the system to undermine fair competition or mislead buyers.
    
    \item \textbf{Black-Box Transferable Evasion/Misclassification Attack (Using Surrogate Data)} (\textbf{Score:} 5.593) \\
    \textbf{Objective:} Integrity \\
    \textbf{Explanation:} Without direct access to the training data, the attacker uses publicly available or existing reviews on the platform to train a substitute model that tries to mimic the feedback scoring model's behavior. Using this model, the attacker generates adversarial reviews optimized for high scores when processed by the scoring model, and submits the reviews to the platform. Such attacks allow malicious actors to amplify the visibility of biased or deceptive feedback reviews.

    \item \textbf{Black-Box Resource-Latency Attack (Requiring Access to Systems' Metrics)} (\textbf{Score:} 2.849) \\
    \textbf{Objective:} Availability \\
    \textbf{Explanation:} An attacker can repeatedly submit variations of reviews and measure the time from when each review was submitted to the platform until the review appears on the platform, in order to craft reviews that intentionally slow down the system's processing time, causing delays in the platform's review display functionality.

    \item \textbf{Black-Box Transferable Resource-Latency Attack (Using Surrogate Data)} (\textbf{Score:} 2.737 )\\
    \textbf{Objective:} Availability \\
    \textbf{Explanation:} Without direct access to the training data, the attacker uses publicly available or existing reviews on the platform to train a substitute model. Using this model, the attacker generates adversarial reviews optimized to increase the model's inference time. These reviews disrupt the platform's operations, causing delays in displaying new reviews for products.

    \item \textbf{Black-Box Interactive Decision-Based Model Extraction Attack (Using Surrogate Data)} (\textbf{Score:} 2.682) \\
    \textbf{Objective:} Privacy \\
    \textbf{Explanation:} The attacker uses surrogate data to interact with the model, analyzing the feedback provided by the system to extract its underlying behavior. By submitting strategically crafted queries and following the system's decisions (examining where reviews are displayed in the review list), the attacker can reconstruct a model that closely mimics the original scoring model. This model can then be used to predict review scores or develop further attacks.

\end{enumerate}

\noindent\textbf{Evaluation Results: Feedback from System Owners and Experts}

\noindent\textbf{System Owners' Feedback:}
The system owners expressed strong agreement with FRAME’s ranking, emphasizing that integrity attacks pose the greatest risk due to their potential to undermine the credibility of the platform. Manipulations in feedback scoring could mislead buyers and harm sellers, directly impacting the platform’s reputation. While availability attacks were acknowledged, the system owners considered them less serious, noting that such issues are likely to be noticed and addressed quickly by platform users or administrators. The ranking aligns well with these priorities, with integrity-focused attacks receiving higher scores than privacy or availability attacks.

\noindent\textbf{Experts' Feedback:}
The experts rated the \emph{overall framework accuracy} at \textbf{9/10} and expressed \emph{strong agreement} with the relevance of the top-ranked attacks, highlighting the alignment between the identified threats and real-world scenarios. 
The experts also confirmed that all critical objectives—\textit{integrity}, \textit{privacy}, and \textit{availability}—were adequately addressed by the framework.
On average, the \emph{attack-specific accuracy} and \emph{attack-specific relevance} metrics for the top-five attacks scored \textbf{8.8/10} and \textbf{8.4/10}, respectively, reflecting FRAME’s strong alignment with real-world threats. 
Integrity-focused attacks, such as the \emph{black-box interactive decision-based evasion/misclassification attack}, stood out with the highest scores among the top five and were rated \textbf{9-10/10} by the experts for both accuracy and relevance, highlighting their critical impact on the system's credibility and the framework’s precision in identifying these as the most serious risks.
Availability and privacy attacks, which were ranked lower by the framework due to their limited impact or detectability, were confirmed by the experts as appropriately scored, affirming the framework's ability to differentiate risks effectively across security objectives.

\subsection{Demonstration: Expanding FRAME with a Countermeasure}
While FRAME’s core functionality focuses on quantifying adversarial risks based on system characteristics, it can be naturally extended to support risk reassessment after the introduction of common countermeasures in adversarial machine learning such as adversarial retraining~\cite{sheshadri2024targeted,tramer2017ensemble} and input sanitation~\cite{ho2022data,smith2025adversarial}. 

We demonstrate this capability using \textit{adversarial retraining} as a representative countermeasure, showing how FRAME can be reapplied to reflect the post-countermeasure threats. To support this demonstration, we employed our LLM-assisted data extraction pipeline (described in Section~\ref{sec:Dataset Creation and Analysis}) to collect academic publications that evaluate adversarial retraining as a defense against various attack techniques. These sources are filtered based on their relevance to the evaluated system (e.g., model architecture, domain) to extract context-specific success rates of attacks under retraining.

Each attack in our attack-feasibility-impact mapping is annotated with a binary indicator specifying whether adversarial retraining is known to mitigate it~\cite{yoo2021towards,omar2022making}. When retraining is applied, FRAME adjusts the risk scores of these mitigated attacks by incorporating the empirically observed post-retraining success rate. Specifically, for a given attack \(a\), its original risk score \(S(a)\) is updated to:
\[
S'(a) = S(a) \times SR_{\text{retrain}}(a),
\]
where \(SR_{\text{retrain}}(a)\) is the success rate of the attack as reported in retraining scenarios. For instance, if an attack originally had a score of \(S(a) = 6.0\) and retraining reduces its success rate to 30\%, the updated score becomes \(S'(a) = 6.0 \times 0.3 = 1.8\).

\begin{figure*}[h]
    \centering
    \includegraphics[width=\linewidth]{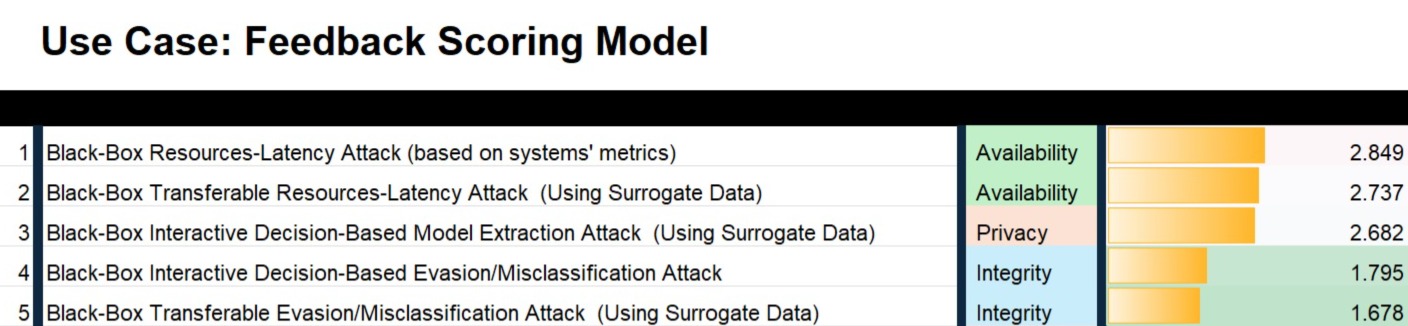}
    \caption{Top-5 ranked threats for the Feedback Scoring Model after applying adversarial retraining.}
    \label{fig:retraining-impact}
\end{figure*}

Figure~\ref{fig:retraining-impact} illustrates this process for the Feedback Scoring Model use case. After applying adversarial retraining, the risk scores of inference-time evasion attacks significantly decreased from 5.984 and 5.593 to 1.795 and 1.678, respectively. In contrast, the scores and relative rankings of other attack categories, such as availability and privacy threats, remained unchanged, as adversarial retraining does not directly mitigate those vectors. This result highlights how FRAME can capture the selective effectiveness of targeted countermeasures and update the threat prioritization accordingly.

\subsection{Summary and Discussion of Evaluation Results Across Use Cases}

Our evaluation of the six use cases demonstrated the framework’s ability to consistently identify and rank adversarial threats effectively, providing actionable insights tailored to each system's unique context.

\textbf{Overall agreement and scoring.}
Across all use cases, both system owners and experts consistently agreed with FRAME's rankings, affirming their alignment with real-world risk levels. The average \emph{overall framework accuracy} across use cases was rated at \textbf{9/10}, with experts providing \textbf{strongly agree} or \textbf{agree} ratings for \emph{top threat relevance} in each of the use cases.

\textbf{Attack-specific accuracy and relevance.}
The average scores for \emph{attack-specific accuracy} and \emph{attack-specific relevance} were impressively high at \textbf{9.2/10} and \textbf{8.9/10} respectively across the evaluated use cases.
Notably, in almost all cases, the attacks identified as most severe by FRAME, consistently received the highest scores from experts for both accuracy and relevance.
This independent agreement by the experts reflects strong confidence in the severity of these attacks, affirming that they are appropriately prioritized.

\textbf{Robust coverage of security objectives.} 
In the majority of use cases, experts confirmed that the framework effectively addressed all critical objectives—\emph{integrity}, \emph{privacy}, and \emph{availability}, with scores proportional to their importance in real-world scenarios. 
Integrity and availability were often prioritized in systems where functionality and reliability were critical (e.g., \emph{traffic steering in O-RAN}), while privacy was elevated in contexts where regulatory compliance was a key concern (e.g., \emph{malware classification for e-mail security}).

\textbf{Consistency in criticality scoring.}
FRAME demonstrated its ability to not only to identify the most critical threats but also to assign scores that reflect the severity of risks and their practical applicability. Experts and system owners consistently validated this nuanced scoring.
In cases with multiple relevant serious threats, such as \emph{beam hopping in satellite communication}, all top-ranked attacks received relatively high scores (\textbf{7.389}, \textbf{6.41}, \textbf{6.35}). 
This reflects the nature of real-time resource optimization systems where both service availability and computational integrity are crucial for operation. Interestingly, while most systems showed similar patterns in their threat priorities, an unexpected emphasis on privacy risks was seen in the \emph{malware cassification for e-mail security} system. When investigating this pattern, we found that regulatory compliance requirements and the sensitive nature of email data made privacy breaches particularly concerning in this context. This contrasts with systems like the \emph{Feedback Scoring Model}, where privacy risks received significantly lower scores (e.g., \textbf{2.682}) compared to integrity-focused attacks (\textbf{5.984}), reflecting their reduced practical impact in that environment.

\textbf{Most sensitive areas and attacks across use cases.}
Across all use cases, integrity consistently emerged as the most critical area, with attacks targeting system credibility and functionality receiving the highest scores. 
For example, \emph{interactive decision-based evasion/misclassification attacks} were ranked among the top threats in nearly every use case, with perfect scores for accuracy and relevance (\textbf{10/10}) in use cases like \emph{traffic steering in O-RAN} and \emph{feedback scoring model}. 
Availability attacks were ranked highly in systems where real-time performance is crucial, such as \emph{beam hopping in satellite communication}, highlighting the importance of system responsiveness and reliability.

\textbf{Point for improvement.}
Following input from the AML experts, the following refinement to the system profiling questionnaire is suggested: adding a specific question to account for query volume. This adjustment would help address cases where attacks involving a high number of queries may be less feasible due to the presence of monitoring systems that increase detectability.

\textbf{Conclusion.}
The evaluation performed highlights FRAME’s ability to not only rank threats effectively but also assign scores that capture the nuanced interplay of attacks' impact, feasibility, and relevance across diverse use cases. 
Its performance in balancing critical objectives and adapting to system-specific requirements highlights its potential as a robust tool for prioritizing adversarial risks and guiding mitigation efforts.

\section{Conclusion and Future Work}
This paper introduces FRAME, a systematic framework that addresses the critical need for comprehensive adversarial risk assessment in ML systems. 
By integrating structured system profiling, expert-driven attack mapping, and empirical data, FRAME delivers actionable and transparent risk evaluations tailored to the unique characteristics of a given system. 
By identifying and prioritizing the most severe threats, the framework equips system owners with the insights needed to make informed decisions and allocate resources effectively. 
This work provides a valuable tool for improving the security of ML systems and lays the groundwork for future research on adversarial risk mitigation strategies.

\noindent\textbf{Countermeasures.}  
while FRAME was shown to be effective in identifying and prioritizing adversarial risks, providing tailored countermeasures strategies is beyond the scope of this paper. Countermeasures could theoretically be generated by providing the identified risks to an LLM tasked with suggesting generic countermeasures. However, such countermeasures would not be tailored to the specific characteristics of the evaluated systems, including their operational constraints, and would not consider the potential impact of the mitigation mechanisms on system performance. Developing a methodology to deliver system-specific and impact-aware countermeasures is a planned direction for future work.

\noindent\textbf{Application to complex systems.}  
with minor adjustments, FRAME is adapted to systems composed of multiple ML models. The questionnaire can be completed for each model individually, capturing model-specific details and system-specific requirements. 
FRAME would then evaluate and rank the top risks for each model separately. 
These risks could be combined in a unified list, ordered by severity, to provide system owners with a clear and comprehensive  understanding of which models and components require the most attention and resources for defense.

\newpage

\section*{Ethical Considerations}
This research was conducted in accordance with ethical standards and poses no foreseeable risks. 
All representatives for the evaluated systems were fully informed about the nature and objectives of this research. To maintain confidentiality, the company name, specific questionnaire responses, and detailed system characteristics are excluded from publication.

\section*{Acknowledgments}
We thank the organizations and individuals who participated in our research. Their insights, feedback, and willingness to share their experiences were essential to this work.

\newpage
\bibliographystyle{elsarticle-num.bst} 
\bibliography{ref.bib}

\clearpage
\section*{Appendix}
\appendix

\section{\label{sec:background}Adversarial Machine Learning}
Adversarial machine learning (AML) refers to the study of the attacks and defenses on ML algorithms~\cite{vassilev2024aml}.
With the increasing deployment and popularity of ML in critical domains, AML has become an essential area of research, addressing vulnerabilities that arise throughout the ML lifecycle.
AML attacks can be categorized based on their objectives, stages, specificity, and attacker knowledge and by the execution mode.

\noindent\textbf{By Objective:}

\begin{itemize}
    \item \textbf{Integrity Attacks:} These attacks aim to cause incorrect model predictions or classifications. Examples include evasion attacks during inference, which craft inputs to deceive the model~\cite{thys2019fooling,shafahi2018poison}.
    \item \textbf{Availability Attacks:} These attacks aim to disrupt the functionality or efficiency of ML systems and include resource-latency attacks that inflate inference times or degrade system performance~\cite{shumailov2021sponge,shapira2023phantom,zhao2022clpa}.
    \item \textbf{Privacy Attacks:} These attacks target sensitive information of the model or training data and include the following attacks: membership inference, model extraction, and  data reconstruction~\cite{shokri2017membership,fredrikson2015model}.
\end{itemize}

\noindent\textbf{By Attack Phase:}
\begin{itemize}
    \item \textbf{Causative Attacks:} 
    In these attacks, adversaries manipulate the training data or the model itself to compromise its functionality. Examples include model poisoning attacks, where the model's parameters are altered~\cite{bagdasaryan2020backdoor,sun2019can}, and data poisoning attacks, in which the training dataset is modified~\cite{gu2019badnets,shafahi2018poison}. Data poisoning can target both data samples and their corresponding labels or focus solely on modifying the data without changing its labels (clean-label attacks)~\cite{gu2019badnets,shafahi2018poison,zhao2022clpa}.
    \item \textbf{Exploratory Attacks:} 
    In these attacks, adversaries manipulate input to deployed models to achieve their objectives without altering the model itself. Examples include membership inference attacks (MIA), which aim to determine if specific data points were part of the training dataset~\cite{shokri2017membership}, and adversarial examples, where inputs are crafted to mislead the model, in order to increase computational costs or  reduce the system's efficiency (e.g., resource-latency attacks~\cite{thys2019fooling,shumailov2021sponge,shapira2023phantom}).
\end{itemize}

\noindent\textbf{By Specificity:}
\begin{itemize}
    \item \textbf{Selective Attacks:} 
    These attacks target specific instances or groups of instances. For example, MIAs aim to determine if a specific instance was part of the training data~\cite{shokri2017membership,leino2020stolen}. Backdoor and targeted poisoning attacks also fall into this category. Backdoor attacks embed a trigger in the training data that causes the model to behave maliciously when the trigger is present during inference. Similarly, targeted poisoning attacks degrade the model’s performance on specific inputs, such as a particular class of data. These techniques focus on specific, preselected objectives~\cite{gu2019badnets,shafahi2018poison,suciu2018does}.
    \item \textbf{Indiscriminate Attacks:} 
    These attacks aim to degrade the overall performance of the model rather than targeting specific instances. Examples include untargeted poisoning attacks that reduce the model’s accuracy across all inputs~\cite{zhao2022clpa,nelson2008exploiting} and model extraction attacks, which replicate the target model’s functionality~\cite{chandrasekaran2020exploring,jagielski2020high}.
\end{itemize}

\noindent\textbf{By Attacker Knowledge (Threat Models):}

\begin{itemize}
    \item \textbf{White-Box Attacks:} 
    In this scenario, the attacker has full access to the target model, including its architecture, parameters, and training data. This comprehensive knowledge enables precise and highly effective attacks, but that require more knowledge. For example, gradient-based adversarial examples, where the attacker leverages gradients to craft inputs that deceive the model~\cite{szegedy2013intriguing, thys2019fooling}.

    \item \textbf{Black-Box Attacks:} 
    Here, the attacker has no direct access to the model's internal details. In some cases, the attacker interacts with the model by submitting queries and analyzing the responses, which could include soft outputs (e.g., confidence scores) or hard outputs (e.g., decisions). In other cases, attackers exploit measurable external metrics, such as inference time or energy consumption, to craft their attacks~\cite{chen2020hopskipjumpattack, ilyas2018prior, shumailov2021sponge}. Black-box attacks often rely on transferability, where adversarial examples are crafted using a substitute model and then applied to the target model~\cite{liu2016delving, wang2021enhancing}.

    \item \textbf{Gray-Box Attacks:} 
    n this case, the attacker possesses partial knowledge of the target system, obtained, for example, via access to the training data distribution or surrogate datasets. This limited information can significantly enhance black-box attack techniques. For instance, knowing the training data's characteristics allows attackers to create a surrogate model more closely aligned with the target model, increasing the likelihood of a successful transferability attack.
\end{itemize}

\noindent\textbf{By Execution Mode (Digital/Physical)}:
AML attacks can be executed in digital environments, physical settings, or through a combination of both.

\begin{itemize}
    \item \textbf{Digital Attacks:} 
    These attacks are conducted entirely within digital systems, targeting the data or model directly in a virtual environment~\cite{shumailov2021sponge, szegedy2013intriguing}.
    \item \textbf{Physical Attacks:} 
    These involve manipulating input in the physical world to deceive ML systems. A common example is modifying real-world objects, such as altering traffic signs to mislead autonomous vehicles\cite{evtimov2017robust, thys2019fooling}.
    \item \textbf{Hybrid Attacks:}  
    Some attacks can exploit vulnerabilities across both digital and physical environments, allowing adversaries to apply the same attack seamlessly in either context. For example, an adversarial pattern crafted digitally can be applied in the physical world to mislead the system, demonstrating flexibility in execution and broadening the scope of potential impact~\cite{thys2019fooling}.
\end{itemize}

\clearpage

\onecolumn

\section{System Profiling - Questionnaire}

In this section we present the original questionnaire, along with the  \textbf{Use Case Customized} questionnaire, as altered automatically using LLM integration in our system according to the use case description (hereby it is the "Feedback Scoring Model" use case), used for system profiling:
\label{Questionnaire}
\label{QuestionnaireCustomized}

{

\newcommand{\restoreheadrule}{\renewcommand{\headrulewidth}{0.4pt}} %

\definecolor{questionbg}{RGB}{240, 248, 255}
\definecolor{choicebg}{RGB}{230, 230, 230}   
\definecolor{headerbg}{RGB}{100, 149, 237}   
\definecolor{titlefg}{RGB}{255, 255, 255}    

\newcommand{\customrule}{\noindent\rule{\linewidth}{1.2pt}\par}

\newcounter{questionNumber}
\renewcommand{\thequestionNumber}{Q\arabic{questionNumber}}
\newcommand{\origQTitle}{\stepcounter{questionNumber} Q\thequestionNumber}
\newcommand{\custQTitle}{Q\thequestionNumber{} -- Customized}

\newtcolorbox{questionbox}[1]{%
    colback=questionbg,
    colframe=headerbg,
    coltitle=titlefg,
    sharp corners,
    boxrule=1.2pt,
    width=\linewidth,
    before=\vspace{0.3cm},
    after=\vspace{-0.3cm},
    fonttitle=\bfseries,
    title={#1}
}

\newtcolorbox{choicebox}{
    colback=choicebg,
    colframe=black,
    sharp corners,
    boxrule=0.5pt,
    width=\linewidth
}

\newcommand{\checkbox}{\ding{113}} 


\section*{Threat-Based Questions}

Impact T1-I (Instance-Specific Incorrect Output): The model is manipulated to produce wrong outputs for particular inputs chosen by the attacker (e.g., using adversarial examples).\\
\noindent
\begin{minipage}[t]{0.495\linewidth}
\begin{questionbox}{Q1 - General}
How severe would the impact be if an attacker caused the model to provide incorrect outputs for specific inputs they selected?  \\
\textbf{Example:} A fraud detection model incorrectly approves transactions flagged by the attacker.
\end{questionbox}
\end{minipage}
\hfill
\begin{minipage}[t]{0.495\linewidth}
\begin{questionbox}{Q1 - Customized}
How severe would the impact be if an attacker could craft specific product reviews that trick the model into assigning incorrect scores - such as ranking low-quality reviews at the top or hiding helpful ones? \\
\end{questionbox}
\end{minipage}

\vspace{0.3cm}
\begin{choicebox}
\begin{tabularx}{\linewidth}{X X X X X}
    \checkbox Very High & \checkbox High & \checkbox Medium Impact & \checkbox Low & \checkbox Very Low
\end{tabularx}
\end{choicebox}
\customrule

\noindent Impact T2-I (Backdoor Attack): The model is manipulated to produce incorrect outputs only for inputs with a specific, predefined pattern chosen by the attacker. \\
\noindent
\begin{minipage}[t]{0.495\linewidth}
\begin{questionbox}{Q2 - General}
How severe would the impact be if a predefined trigger in the inputs caused the model to produce incorrect outputs?  \\
\textbf{Example:} A specific phrase in a document bypasses a text moderation model.
\end{questionbox}
\end{minipage}
\hfill
\begin{minipage}[t]{0.495\linewidth}
\begin{questionbox}{Q2 - Customized}
How severe would the impact be if a specific hidden trigger (e.g., a keyword, symbol, or formatting pattern) in a review always caused the model to assign an inaccurate score, regardless of the review’s actual quality?
\\
\end{questionbox}
\end{minipage}

\vspace{0.3cm}
\begin{choicebox}
\begin{tabularx}{\linewidth}{X X X X X}
    \checkbox Very High & \checkbox High & \checkbox Medium Impact & \checkbox Low & \checkbox Very Low
\end{tabularx}
\end{choicebox}

\customrule

\noindent Impact T3-I (Targeted Attack): The model's predictions are manipulated to be incorrect for a chosen subset of inputs while the rest remain unaffected.\\
\noindent
\begin{minipage}[t]{0.495\linewidth}
\begin{questionbox}{Q3 - General}
How severe would the impact be if an attacker altered the system’s behavior to produce incorrect outputs for a specific subset of inputs based on their inherent characteristics?  \\
\textbf{Example:} An attacker manipulates a financial model so that transactions from a specific region are consistently flagged as high-risk, regardless of their actual attributes.
\end{questionbox}
\end{minipage}
\hfill
\begin{minipage}[t]{0.495\linewidth}
\begin{questionbox}{Q3 - Customized}
How severe would the impact be if an attacker manipulated the model to consistently mis-score a specific subset of reviews based on inherent traits - such as topic, language style, or reviewer identity - regardless of their actual quality?  \\
\end{questionbox}
\end{minipage}

\vspace{0.3cm}
\begin{choicebox}
\begin{tabularx}{\linewidth}{X X X X X}
    \checkbox Very High & \checkbox High & \checkbox Medium Impact & \checkbox Low & \checkbox Very Low
\end{tabularx}
\end{choicebox}

\customrule

\noindent Impact T4-P (Membership Inference Attack - MIA): Determining if a specific data point was part of the model’s training dataset. \\
\noindent
\begin{minipage}[t]{0.495\linewidth}
\begin{questionbox}{Q4 - General}
How severe would the impact be if an attacker could identify whether specific data samples were used in training?  \\
\textbf{Example:} A malicious actor determines a user’s medical record was part of a training dataset.
\end{questionbox}
\end{minipage}
\hfill
\begin{minipage}[t]{0.495\linewidth}
\begin{questionbox}{Q4 - Customized}
How severe would the impact be if an attacker could determine whether a specific user review was part of the training dataset used to build the scoring model?  \\
\end{questionbox}
\end{minipage}

\vspace{0.3cm}
\begin{choicebox}
\begin{tabularx}{\linewidth}{X X X X X}
    \checkbox Very High & \checkbox High & \checkbox Medium Impact & \checkbox Low & \checkbox Very Low
\end{tabularx}
\end{choicebox}

\customrule

\noindent Impact T5-P (Training Set Attribute Inference): Inferring the characteristics of the dataset used to train the model. \\
\noindent
\begin{minipage}[t]{0.495\linewidth}
\begin{questionbox}{Q5 - General}
How severe would the impact be if an attacker inferred general properties of the training data (e.g., feature distributions, input types)?  \\
\textbf{Example:} An attacker deduces the demographic makeup of a training dataset.
\end{questionbox}
\end{minipage}
\hfill
\begin{minipage}[t]{0.495\linewidth}
\begin{questionbox}{Q5 - Customized}
How severe would the impact be if an attacker could infer general characteristics of the training dataset - such as favored writing styles, sentiment tendencies, or common reviewer behaviors?  \\
\end{questionbox}
\end{minipage}

\vspace{0.3cm}
\begin{choicebox}
\begin{tabularx}{\linewidth}{X X X X X}
    \checkbox Very High & \checkbox High & \checkbox Medium Impact & \checkbox Low & \checkbox Very Low
\end{tabularx}
\end{choicebox}

\customrule

\noindent Impact T6-P (Training Set Reconstruction): Reconstructing data used to train the model. \\
\noindent
\begin{minipage}[t]{0.495\linewidth}
\begin{questionbox}{Q6 - General}
How severe would the impact be if an attacker leaked some or all of the training data?  \\
\textbf{Example:} Customer data used for training a recommendation system is exposed.
\end{questionbox}
\end{minipage}
\hfill
\begin{minipage}[t]{0.495\linewidth}
\begin{questionbox}{Q6 - Customized}
How severe would the impact be if an attacker gained access to some or all of the reviews used to train the scoring model - including those not intended for public display?  \\
\end{questionbox}
\end{minipage}

\vspace{0.3cm}
\begin{choicebox}
\begin{tabularx}{\linewidth}{X X X X X}
    \checkbox Very High & \checkbox High & \checkbox Medium Impact & \checkbox Low & \checkbox Very Low
\end{tabularx}
\end{choicebox}

\customrule

\noindent Impact T7-P (Model Replication Attack): Extract information about the model (usually by querying the model) in order to create a replica of the model. \\
\noindent
\begin{minipage}[t]{0.495\linewidth}
\begin{questionbox}{Q7 - General}
How severe would the impact be if an attacker extracted enough information to replicate your model?  \\
\textbf{Example:} An attacker uses API queries to recreate a model for commercial gain.
\end{questionbox}
\end{minipage}
\hfill
\begin{minipage}[t]{0.495\linewidth}
\begin{questionbox}{Q7 - Customized}
How severe would the impact be if an attacker were able to create a copy of the feedback scoring model (for example, to simulate the score their reviews receive)?  \\
\end{questionbox}
\end{minipage}

\vspace{0.3cm}
\begin{choicebox}
\begin{tabularx}{\linewidth}{X X X X X}
    \checkbox Very High & \checkbox High & \checkbox Medium Impact & \checkbox Low & \checkbox Very Low
\end{tabularx}
\end{choicebox}

\customrule

 \noindent Impact T8-A (Increased Resource Usage for All Inputs): The model consumes more resources for every input during inference, leading to higher operational costs. \\
\noindent
\begin{minipage}[t]{0.495\linewidth}
\begin{questionbox}{Q8 - General}
How severe would the impact be if an attacker caused the ML model to use excessive resources (e.g., energy, time) for all inputs at inference time?  \\
\textbf{Example:} The system's response times slow down across all operations.
\end{questionbox}
\end{minipage}
\hfill
\begin{minipage}[t]{0.495\linewidth}
\begin{questionbox}{Q8 - Customized}
How severe would the impact be if the attacker caused the system to consume excessive compute resources or time for scoring reviews across the entire platform, slowing down review processing for all users?  \\
\end{questionbox}
\end{minipage}

\vspace{0.3cm}
\begin{choicebox}
\begin{tabularx}{\linewidth}{X X X X X}
    \checkbox Very High & \checkbox High & \checkbox Medium Impact & \checkbox Low & \checkbox Very Low
\end{tabularx}
\end{choicebox}

\customrule

\noindent Impact T9-A (Increased Resource Usage for Specific Inputs): The model consumes more resources for particular inputs chosen by the attacker during inference. \\
\noindent
\begin{minipage}[t]{0.495\linewidth}
\begin{questionbox}{Q9 - General}
How severe would the impact be if an attacker caused the ML model to use excessive resources for specific inputs at inference time?  \\
\textbf{Example:} When processing certain inputs, the system experiences disproportionate delays, with higher-than-usual energy or computational requirements, while other inputs are processed normally.
\end{questionbox}
\end{minipage}
\hfill
\begin{minipage}[t]{0.495\linewidth}
\begin{questionbox}{Q9 - Customized}
How severe would the impact be if certain crafted reviews caused unusually high processing time or system resource usage, while other reviews were scored normally? \\
\end{questionbox}
\end{minipage}

\vspace{0.3cm}
\begin{choicebox}
\begin{tabularx}{\linewidth}{X X X X X}
    \checkbox Very High & \checkbox High & \checkbox Medium Impact & \checkbox Low & \checkbox Very Low
\end{tabularx}
\end{choicebox}

\customrule

\noindent Impact T10-A (Increased Resource Usage During Training): The model requires more resources during the training process, extending the training duration or increasing costs. \\
\noindent
\begin{minipage}[t]{0.495\linewidth}
\begin{questionbox}{Q10 - General}
How severe would the impact be if an attacker caused the training process to use excessive resources (e.g., energy consumption, time, etc.)?  \\
\textbf{Example:} The training process becomes inefficient, taking longer and consuming more computational resources than expected.
\end{questionbox}
\end{minipage}
\hfill
\begin{minipage}[t]{0.495\linewidth}
\begin{questionbox}{Q10 - Customized}
How severe would the impact be if an attacker manages to significantly increased the time, cost, or complexity of retraining the feedback scoring model?  \\
\end{questionbox}
\end{minipage}

\vspace{0.3cm}
\begin{choicebox}
\begin{tabularx}{\linewidth}{X X X X X}
    \checkbox Very High & \checkbox High & \checkbox Medium Impact & \checkbox Low & \checkbox Very Low
\end{tabularx}
\end{choicebox}

\customrule

\noindent Impact T11-A (Incorrect Outputs for All Inputs): The model produces incorrect outputs for every input across the board (e.g., using poisoning attacks). \\
\noindent
\begin{minipage}[t]{0.495\linewidth}
\begin{questionbox}{Q11 - General}
How severe would the impact be if an attacker caused the model to produce wrong outputs for every input?  \\
\end{questionbox}
\end{minipage}
\hfill
\begin{minipage}[t]{0.495\linewidth}
\begin{questionbox}{Q11 - Customized}
How severe would the impact be if the model consistently rank reviews incorrectly – showing unhelpful or malicious reviews at the top of the page, while hiding quality reviews? \\
\end{questionbox}
\end{minipage}

\vspace{0.3cm}
\begin{choicebox}
\begin{tabularx}{\linewidth}{X X X X X}
    \checkbox Very High & \checkbox High & \checkbox Medium Impact & \checkbox Low & \checkbox Very Low
\end{tabularx}
\end{choicebox}

\customrule \bigskip

\section*{General Security Questions}
Manipulating the Training Data
\\
\noindent
\begin{minipage}[t]{0.495\linewidth}
\begin{questionbox}{Q12 - General}
How easy is it for the threat actor to compromise the data used for training or retraining the model?
\end{questionbox}
\end{minipage}
\hfill
\begin{minipage}[t]{0.495\linewidth}
\begin{questionbox}{Q12 - Customized}
How easy is it for the threat actor to inject or manipulate reviews that end up in the training dataset of the feedback scoring model?
\end{questionbox}
\end{minipage}

\vspace{0.3cm}
\begin{choicebox}
\begin{tabularx}{\linewidth}{X X X X X}
    \checkbox Very Easy & \checkbox Easy & \checkbox Medium & \checkbox Hard & \checkbox Very Hard
\end{tabularx}
\end{choicebox}

\customrule

Label Manipulation (Label Access)
\\
\noindent
\begin{minipage}[t]{0.495\linewidth}
\begin{questionbox}{Q13 - General}
How easy is it for the threat actor to compromise the labeling procedure (i.e., causing incorrect labels to be included in the training set)?
\end{questionbox}
\end{minipage}
\hfill
\begin{minipage}[t]{0.495\linewidth}
\begin{questionbox}{Q13 - Customized}
How easy is it for the threat actor to influence how reviews are labeled during model training? (For example, providing feedback if the reviews are helpful.)
\end{questionbox}
\end{minipage}

\vspace{0.3cm}
\begin{choicebox}
\begin{tabularx}{\linewidth}{X X X X X}
    \checkbox Very Easy & \checkbox Easy & \checkbox Medium & \checkbox Hard & \checkbox Very Hard
\end{tabularx}
\end{choicebox}

\customrule

Feature Engineering \\
\noindent
\begin{minipage}[t]{0.495\linewidth}
\begin{questionbox}{Q14 - General}
Do you perform any type of manual feature engineering?
\end{questionbox}
\end{minipage}
\hfill
\begin{minipage}[t]{0.495\linewidth}
\begin{questionbox}{Q14 - Customized}
Do you manually define any features from the reviews — such as length, sentiment score, keyword frequency, or readability?\end{questionbox}
\end{minipage}

\vspace{0.3cm}
\begin{choicebox}
\begin{tabular}{p{4 cm} p{3.5cm} p{4cm}}
    \checkbox No manual feature engineering & \checkbox Basic manual feature engineering & \checkbox Extensive manual feature engineering
\end{tabular}
\end{choicebox}

\customrule

 
Performing Training Data Validation
\\
\noindent
\begin{minipage}[t]{0.495\linewidth}
\begin{questionbox}{Q15 - General}
What level of training data validation (e.g., cleaning, filtering, OOD/anomaly detection) is performed?
\end{questionbox}
\end{minipage}
\hfill
\begin{minipage}[t]{0.495\linewidth}
\begin{questionbox}{Q15 - Customized}
What level of validation or filtering is applied to the reviews used for training?
(For example, checking for spammy reviews, or containing external links.)
\end{questionbox}
\end{minipage}

\vspace{0.3cm}
\begin{choicebox}
\begin{tabularx}{\linewidth}{X X X}
  \checkbox None & \checkbox Basic & \checkbox Extensive
\end{tabularx}
\end{choicebox}

\customrule

 
Performing Testing-Data Validation\\
\noindent
\begin{minipage}[t]{0.495\linewidth}
\begin{questionbox}{Q16 - General}
What level of data validation (e.g., cleaning, filtering, OOD/anomaly detection) is performed?  \\
(At inference - when the system is already deployed)
\end{questionbox}
\end{minipage}
\hfill
\begin{minipage}[t]{0.495\linewidth}
\begin{questionbox}{Q16 - Customized}
What level of validation is performed on reviews at inference time (i.e., before they are scored and displayed)?
\end{questionbox}
\end{minipage}

\vspace{0.3cm}
\begin{choicebox}
\begin{tabularx}{\linewidth}{X X X}
  \checkbox None & \checkbox Basic & \checkbox Extensive
\end{tabularx}
\end{choicebox}

\customrule

 
Manipulating the Model's Parameters \\
\noindent
\begin{minipage}[t]{0.495\linewidth}
\begin{questionbox}{Q17 - General}
How secure are your model’s parameters against unauthorized changes?  \\
(e.g., sending malicious updates)
\end{questionbox}
\end{minipage}
\hfill
\begin{minipage}[t]{0.495\linewidth}
\begin{questionbox}{Q17 - Customized}
How secure are the feedback scoring model’s parameters against unauthorized updates that could affect how reviews are scored?
\end{questionbox}
\end{minipage}

\vspace{0.3cm}
\begin{choicebox}
\begin{tabularx}{\linewidth}{X X X}
    \checkbox Very Insecure & \checkbox Insecure & \checkbox Moderately Secure \\
    \checkbox Secure & \checkbox Very Secure &
\end{tabularx}
\end{choicebox}

\customrule

 
Model Retraining Frequency \\
\noindent
\begin{minipage}[t]{0.495\linewidth}
\begin{questionbox}{Q18 - General}
How often do you retrain the model?
\end{questionbox}
\end{minipage}
\hfill
\begin{minipage}[t]{0.495\linewidth}
\begin{questionbox}{Q18 - Customized}
How frequently is the feedback scoring model retrained using new or updated reviews?
\end{questionbox}
\end{minipage}

\vspace{0.3cm}
\begin{choicebox}
\begin{tabularx}{\linewidth}{X X X X X}
    \checkbox Always & \checkbox Often & \checkbox Sometimes & \checkbox Rarely & \checkbox Very Rarely
\end{tabularx}
\end{choicebox}

\customrule

 
Model Feedback (Training) \\
\noindent
\begin{minipage}[t]{0.495\linewidth}
\begin{questionbox}{Q19 - General}
What feedback is accessible to the threat actor during the model's training process?  \\
(Feedback refers to any information the threat actor could potentially access about the model's behavior or decisions during training.)
\end{questionbox}
\end{minipage}
\hfill
\begin{minipage}[t]{0.495\linewidth}
\begin{questionbox}{Q19 - Customized}
What kind of internal feedback might a threat actor gain access to during the model’s training process? \\
(e.g., they see the review’s score, notice if it’s shown or hidden, access internal training logic, or see nothing at all)
\end{questionbox}
\end{minipage}

\vspace{0.3cm}
\begin{choicebox}
\begin{tabular}{p{7cm} p{7cm}}
    \checkbox Full access to the model's flow & \checkbox Decision-based \\
    \checkbox Score-based & \checkbox No feedback
\end{tabular}
\end{choicebox}

\customrule

 
End-to-End/Transfer Learning \\
\noindent
\begin{minipage}[t]{0.495\linewidth}
\begin{questionbox}{Q20 - General}
When retraining the model, do you retrain it from scratch or perform fine-tuning or transfer learning?
\end{questionbox}
\end{minipage}
\hfill
\begin{minipage}[t]{0.495\linewidth}
\begin{questionbox}{Q20 - Customized}
When retraining the feedback scoring model with new reviews, do you fine-tune the existing model or retrain it from scratch?
\end{questionbox}
\end{minipage}

\vspace{0.3cm}
\begin{choicebox}
\begin{tabularx}{\linewidth}{X X}
    \checkbox Fine Tuning & \checkbox Train From Scratch
\end{tabularx}
\end{choicebox}

\customrule

 
Model Online Evaluation\\
\noindent
\begin{minipage}[t]{0.495\linewidth}
\begin{questionbox}{Q21 - General}
Do you conduct online evaluation of your models? Does this include A/B testing?
\end{questionbox}
\end{minipage}
\hfill
\begin{minipage}[t]{0.495\linewidth}
\begin{questionbox}{Q21 - Customized}
Do you conduct online experiments like A/B testing to compare different feedback scoring models in production?
\end{questionbox}
\end{minipage}

\vspace{0.3cm}
\begin{choicebox}
\begin{tabularx}{\linewidth}{X X X}
    \checkbox No Evaluation & \checkbox Yes, but without A/B testing & \checkbox Yes, with A/B testing
\end{tabularx}
\end{choicebox}

\customrule

 
Trigger the Attack - Digital\\
\noindent
\begin{minipage}[t]{0.495\linewidth}
\begin{questionbox}{Q22 - General}
How easy is it for the threat actor to manipulate the model’s inputs digitally at serving time?  \\
(e.g., via an accessible API)
\end{questionbox}
\end{minipage}
\hfill
\begin{minipage}[t]{0.495\linewidth}
\begin{questionbox}{Q22 - Customized}
How easy is it for the threat actor to submit new product reviews that are processed by the scoring model?
\end{questionbox}
\end{minipage}

\vspace{0.3cm}
\begin{choicebox}
\begin{tabularx}{\linewidth}{X X X X X X}
    \checkbox Very Easy & \checkbox Easy & \checkbox Medium & \checkbox Hard & \checkbox Very Hard  & \checkbox Not Possible
\end{tabularx}
\end{choicebox}

\customrule

 
Manipulating the Environment (Trigger the Attack - Physical)\\
\noindent
\begin{minipage}[t]{0.495\linewidth}
\begin{questionbox}{Q23 - General}
How easy is it for the threat actor to manipulate the model’s inputs physically at serving time?  \\
(e.g., placing an adversarial object in front of a sensor)
\end{questionbox}
\end{minipage}
\hfill
\begin{minipage}[t]{0.495\linewidth}
\begin{questionbox}{Q23 - Customized}
Is there any realistic way for a threat actor to manipulate the feedback scoring model through physical means? 
(e.g., by physically altering the review interface, the user's environment, or content presentation in a way that influences how the model scores reviews) \\
Note: In most online platforms like this, physical manipulation is typically not applicable - you may consider marking “Not Possible.”
\end{questionbox}
\end{minipage}

\vspace{0.3cm}
\begin{choicebox}
\begin{tabularx}{\linewidth}{X X X X X X}
    \checkbox Very Easy & \checkbox Easy & \checkbox Medium & \checkbox Hard & \checkbox Very Hard & \checkbox Not Possible
\end{tabularx}
\end{choicebox}

\customrule

 
Model Feedback - Serving\\
\noindent
\begin{minipage}[t]{0.495\linewidth}
\begin{questionbox}{Q24 - General}
What is the feedback provided to the model's users at serving time?
\end{questionbox}
\end{minipage}
\hfill
\begin{minipage}[t]{0.495\linewidth}
\begin{questionbox}{Q24 - Customized}
What type of feedback is provided to users after submitting a review?
e.g., Do users see the direct output of the feedback scoring model (score-based), can infer the review’s score from its position in the review list (decision-based), have access to internal model logic—full access, or receive no feedback at all?
\end{questionbox}
\end{minipage}

\vspace{0.3cm}
\begin{choicebox}
\begin{tabular}{p{7cm} p{7cm}}
    \checkbox Full access to the model's flow & \checkbox Decision-based \\
    \checkbox Score-based & \checkbox No feedback
\end{tabular}
\end{choicebox}

\customrule

 
Knowledge of Training Data\\
\noindent
\begin{minipage}[t]{0.495\linewidth}
\begin{questionbox}{Q25 - General}
How easy is it for a threat actor to know the exact dataset used to train your model?  \\
(e.g., through public documentation, disclosures in research papers, or inference from model behavior)
\end{questionbox}
\end{minipage}
\hfill
\begin{minipage}[t]{0.495\linewidth}
\begin{questionbox}{Q25 - Customized}
How easy is it for a threat actor to find out exactly which product reviews were used to train the scoring model — for example, through public platform data, documentation, or by observing model behavior?
\end{questionbox}
\end{minipage}

\vspace{0.3cm}
\begin{choicebox}
\begin{tabularx}{\linewidth}{X X X X X}
    \checkbox Very Easy & \checkbox Easy & \checkbox Medium & \checkbox Hard & \checkbox Very Hard
\end{tabularx}
\end{choicebox}

\customrule

 
Knowledge: Surrogate Data\\
\noindent
\begin{minipage}[t]{0.495\linewidth}
\begin{questionbox}{Q26 - General}
How easy is it for a threat actor to find or create a dataset similar to the one used to train your model (a surrogate dataset)?  \\
(e.g., by using public datasets, scraping data from similar sources, or collecting data themselves to approximate the training data)
\end{questionbox}
\end{minipage}
\hfill
\begin{minipage}[t]{0.495\linewidth}
\begin{questionbox}{Q26 - Customized}
How easy is it for a threat actor to collect or create a dataset similar to your training data \\
(e.g., by scraping public reviews from the platform)?
\end{questionbox}
\end{minipage}

\vspace{0.3cm}
\begin{choicebox}
\begin{tabularx}{\linewidth}{X X X X X}
    \checkbox Very Easy & \checkbox Easy & \checkbox Medium & \checkbox Hard & \checkbox Very Hard
\end{tabularx}
\end{choicebox}

\customrule

 
Knowledge: Model Data\\
\noindent
\begin{minipage}[t]{0.495\linewidth}
\begin{questionbox}{Q27 - General}
How much information does the threat actor have about the model’s design, training process, or deployment setup?
\end{questionbox}
\end{minipage}
\hfill
\begin{minipage}[t]{0.495\linewidth}
\begin{questionbox}{Q27 - Customized}
How much does the threat actor know about the model architecture or how the review scores are computed?
\end{questionbox}
\end{minipage}

\vspace{0.3cm}
\begin{choicebox}
\begin{tabular}{p{5cm} p{5cm} p{5cm}}
    \checkbox Complete Knowledge & \checkbox Known architecture & \checkbox Task \\
    \checkbox Algorithm & \checkbox Hyperparameters & \checkbox Unknown
\end{tabular}
\end{choicebox}

\customrule

 
Access to Systems' Metrics - Training\\
\noindent
\begin{minipage}[t]{0.495\linewidth}
\begin{questionbox}{Q28 - General}
How easy is it for the threat actor to access systems' metrics in training time?  \\
(e.g., time, energy consumption, CPU)
\end{questionbox}
\end{minipage}
\hfill
\begin{minipage}[t]{0.495\linewidth}
\begin{questionbox}{Q28 - Customized}
How easy is it for a threat actor to monitor system metrics during training (e.g., compute usage or processing time when retraining with new reviews)?
\end{questionbox}
\end{minipage}

\vspace{0.3cm}
\begin{choicebox}
\begin{tabularx}{\linewidth}{X X X X X}
    \checkbox Very Easy & \checkbox Easy & \checkbox Medium & \checkbox Hard & \checkbox Very Hard
\end{tabularx}
\end{choicebox}

\customrule

 
Access to Systems' Metrics - Serving\\
\noindent
\begin{minipage}[t]{0.495\linewidth}
\begin{questionbox}{Q29 - General}
How easy is it for the threat actor to access systems' metrics in serving time?  \\
(e.g., time, energy consumption, CPU)
\end{questionbox}
\end{minipage}
\hfill
\begin{minipage}[t]{0.495\linewidth}
\begin{questionbox}{Q29 - Customized}
How easy is it for a threat actor to monitor system metrics during review scoring (e.g., delays in review display)?
\end{questionbox}
\end{minipage}

\vspace{0.3cm}
\begin{choicebox}
\begin{tabularx}{\linewidth}{X X X X X}
    \checkbox Very Easy & \checkbox Easy & \checkbox Medium & \checkbox Hard & \checkbox Very Hard
\end{tabularx}
\end{choicebox}

\customrule \bigskip

\section*{Use Case Characteristic Questions}


Model Architecture\\
\noindent
\begin{minipage}[t]{0.495\linewidth}
\begin{questionbox}{Q30 - General}
What type of model do you use in the selected use case? \\
\textbf{(Choose the option that best describes your model’s architecture or technique.)}
\end{questionbox}
\end{minipage}
\hfill
\begin{minipage}[t]{0.495\linewidth}
\begin{questionbox}{Q30 - Customized}
What type of model architecture is used to score and rank product reviews on the platform?  
\textbf{(Choose the option that best describes the scoring model’s structure.)}
\end{questionbox}
\end{minipage}

\begin{choicebox}
\begin{tabular}{p{4cm} p{8cm}}
    \checkbox Deep Learning & \checkbox Ensemble \\
    \checkbox Decision Trees & \checkbox Standard ML (e.g., SVM, k-means, etc.)
\end{tabular}
\end{choicebox}

\customrule



Model Task\\
\noindent
\begin{minipage}[t]{0.495\linewidth}
\begin{questionbox}{Q31 - General}
What type of task is your model solving?  \\
\textbf{(Indicates the primary task the model is designed for.)}
\end{questionbox}
\end{minipage}
\hfill
\begin{minipage}[t]{0.495\linewidth}
\begin{questionbox}{Q31 - Customized}
Which machine learning task best describes how the model processes and scores product reviews?
\end{questionbox}
\end{minipage}

\begin{choicebox}
\begin{tabular}{p{4cm} p{4cm} p{4cm}}
    \checkbox Classification & \checkbox Semi-supervised & \checkbox Unsupervised  \\
    \checkbox Regression & \checkbox LLM & \checkbox Object Detection \\
    \multicolumn{3}{l}{\checkbox Reinforcement Learning}
\end{tabular}
\end{choicebox}

\customrule


Data Type\\
\noindent
\begin{minipage}[t]{0.495\linewidth}
\begin{questionbox}{Q32 - General}
What type of data is your model using?  \\
\textbf{(Select the type of data processed by the model.)}
\end{questionbox}
\end{minipage}
\hfill
\begin{minipage}[t]{0.495\linewidth}
\begin{questionbox}{Q32 - Customized}
What type of input data does your model use to evaluate and score product reviews? \\
\end{questionbox}
\end{minipage}
\begin{choicebox}
\begin{tabular}{p{5.5cm} p{5.5cm}}
    \checkbox Images (Computer Vision) & \checkbox Text (NLP) \\
    \checkbox Tabular & \checkbox Voice
\end{tabular}
\end{choicebox}

\customrule \bigskip
 

Use Case Domain \\
\noindent
\begin{minipage}[t]{0.495\linewidth}
\begin{questionbox}{Q33 - General}
What is the domain of your use case?  \\
\textbf{(Identify the application area your model is serving.)}
\end{questionbox}
\end{minipage}
\hfill
\begin{minipage}[t]{0.495\linewidth}
\begin{questionbox}{Q33 - Customized}
What is the main application domain of your feedback scoring model?
\end{questionbox}
\end{minipage}

\vspace{-0.2cm}
\begin{choicebox}
\begin{tabular}{p{3.5cm} p{3.5cm} p{3.5cm}}
    \checkbox Cyber & \checkbox Finance & \checkbox Computer Vision \\
    \checkbox Speech & \checkbox NLP (Text) & \checkbox Network \\
    \multicolumn{3}{l}{\checkbox Recommender System}
\end{tabular}
\end{choicebox}

\customrule \bigskip

}

\vspace{10pt} 

\clearpage
\newpage

\section{Attack Feasibility and Impact Mapping}
The following tables present the defined AML attacks and their mapping to the attacker feasibility factors required for execution (see Table ~\ref{AttCapMap}), followed by a mapping of AML attacks to their targeted impacts (see Table~\ref{Attack impact mapping}).
\label{AttackFeasibilityImpactMapping}
\begin{landscape}
    \begin{figure*}[p]
        \captionsetup{type=table} 
        \label{Attack-Capabilities-Mapping}
        \centering
        \includegraphics[width=0.8\paperheight]{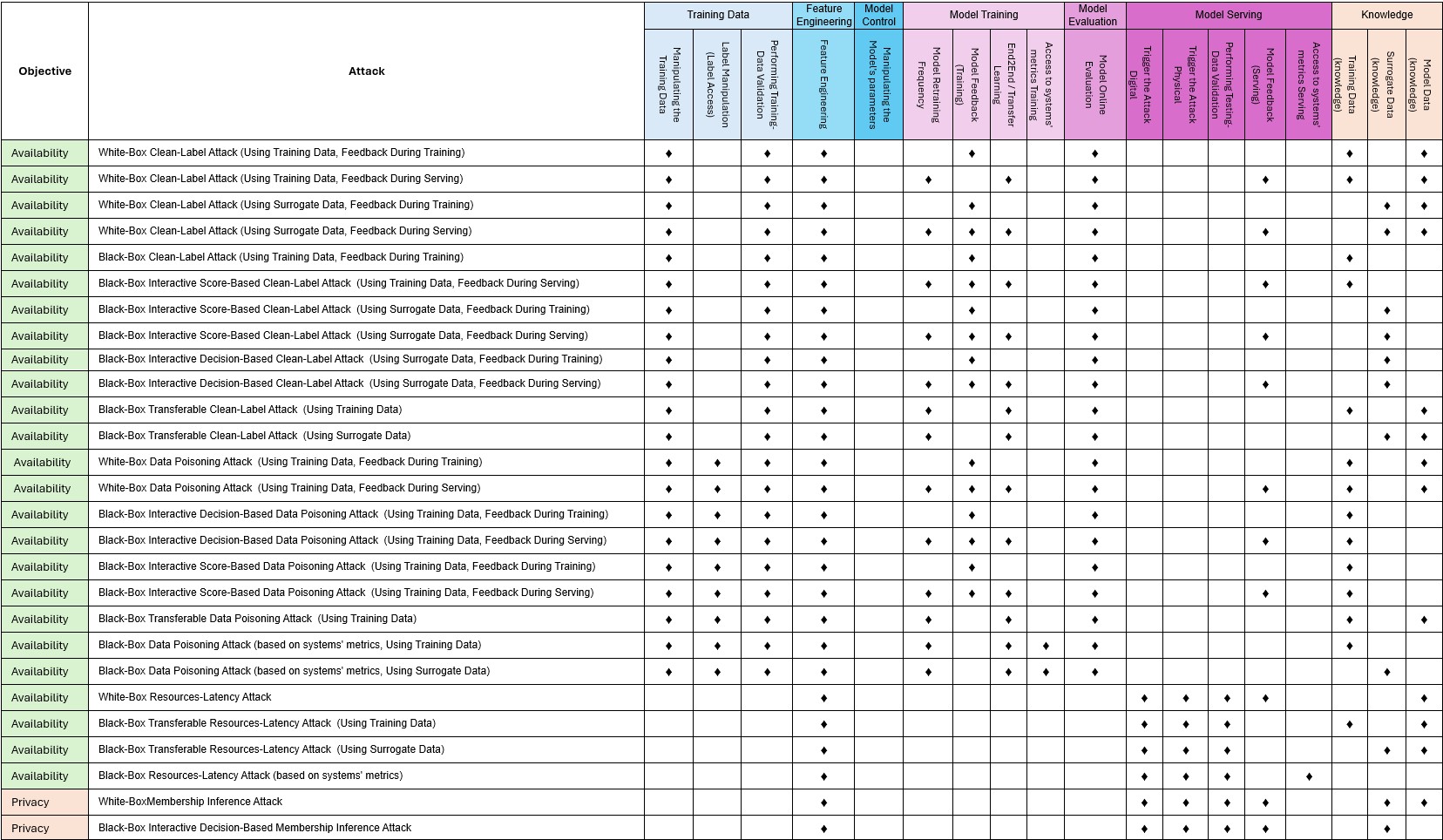}
        \caption{Attack feasibility mapping (1/3).}
        \label{AttCapMap}
    \end{figure*}

    \begin{figure*}[p]
        \captionsetup{type=table} 

        \centering
        \includegraphics[width=0.8\paperheight]{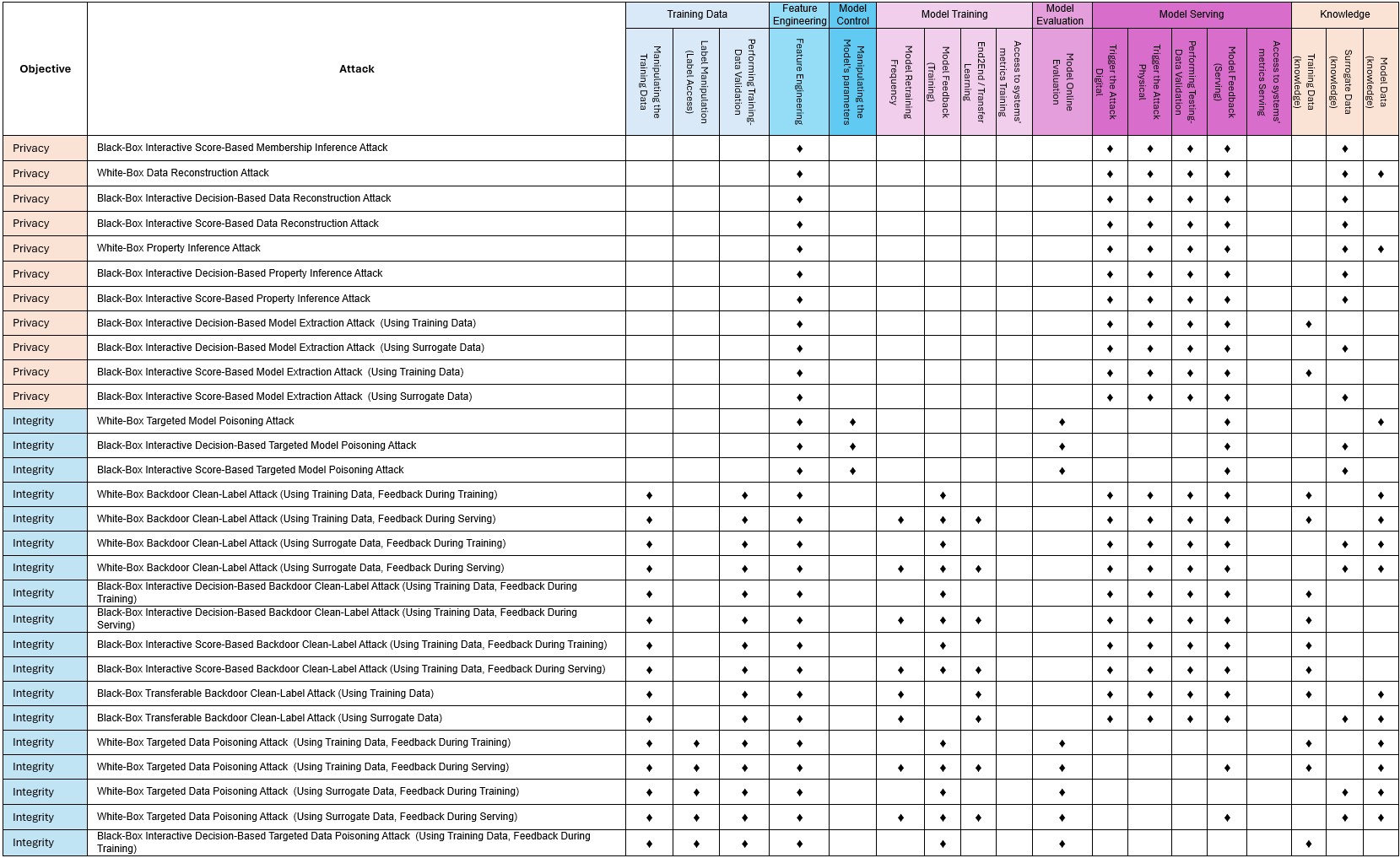}
        \caption{Attack feasibility mapping (2/3).}
    \end{figure*}
    
    \begin{figure*}[p]
        \captionsetup{type=table} 

        \centering
        \includegraphics[width=0.8\paperheight]{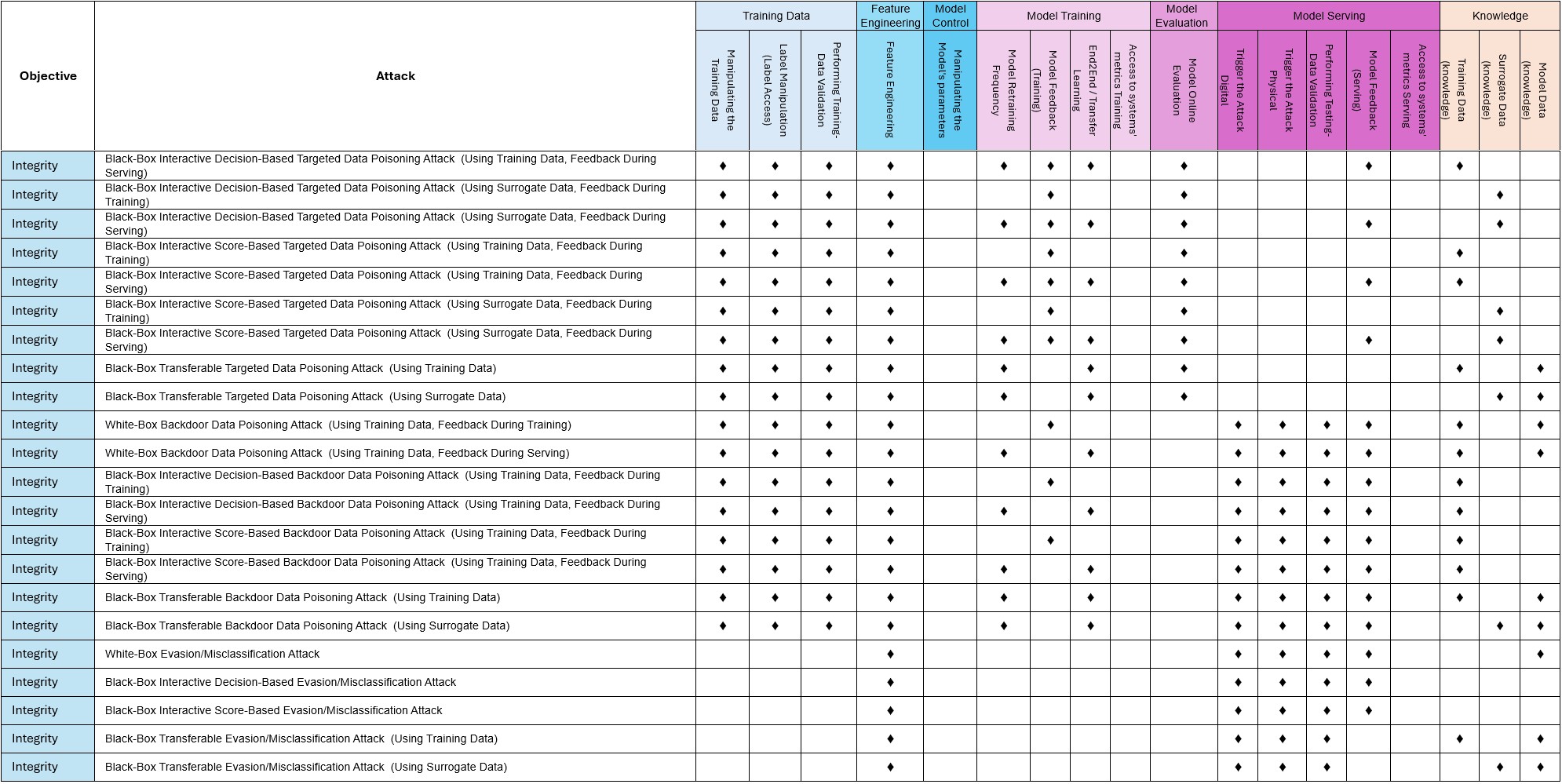}
        \caption{Attack feasibility mapping (3/3).}
    \end{figure*}

\end{landscape}

\begin{landscape}
    \begin{figure*}[p]
        \captionsetup{type=table} 
        \centering
        \includegraphics[width=0.7\paperheight]{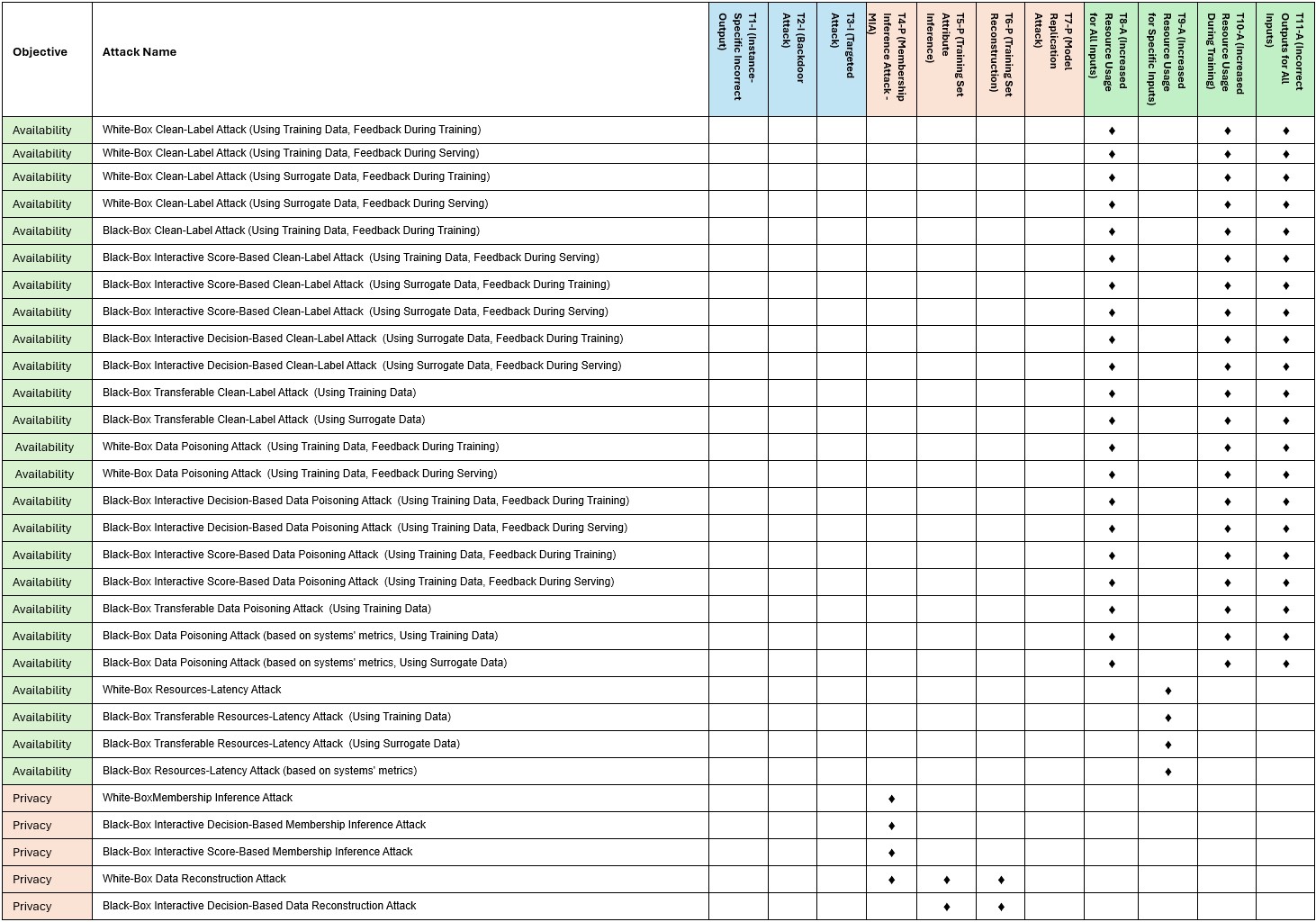}
        \caption{Attack impact mapping (1/3).}
        \label{Attack impact mapping}
        \label{AImp1}
    \end{figure*}

    \begin{figure*}[p]
        \captionsetup{type=table} 
        \centering
        \includegraphics[width=0.7\paperheight]{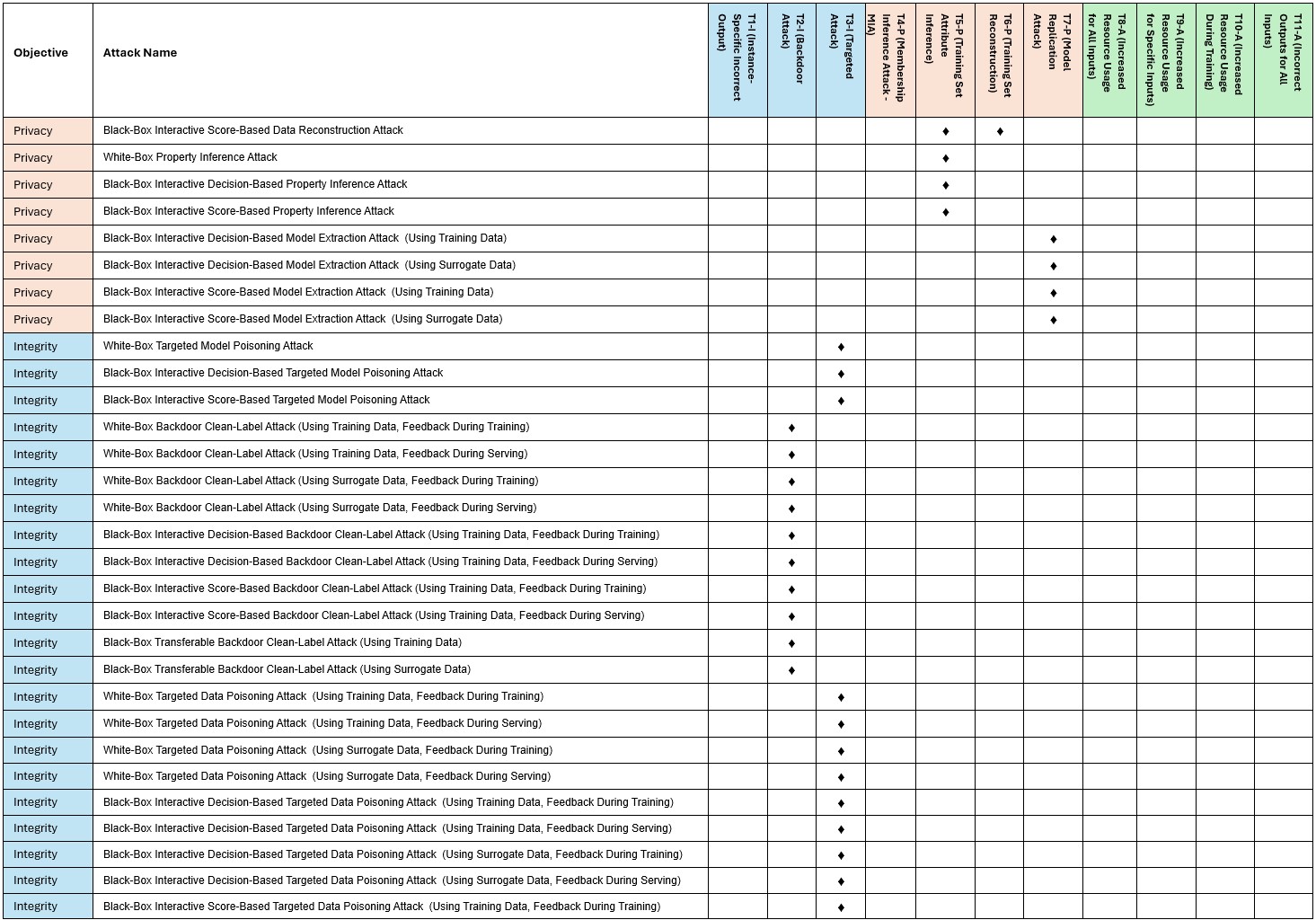}
        \caption{Attack impact mapping (2/3).}
        \label{AImp2}
    \end{figure*}

    \begin{figure*}[p]
        \captionsetup{type=table} 
        \centering
        \includegraphics[width=0.7\paperheight]{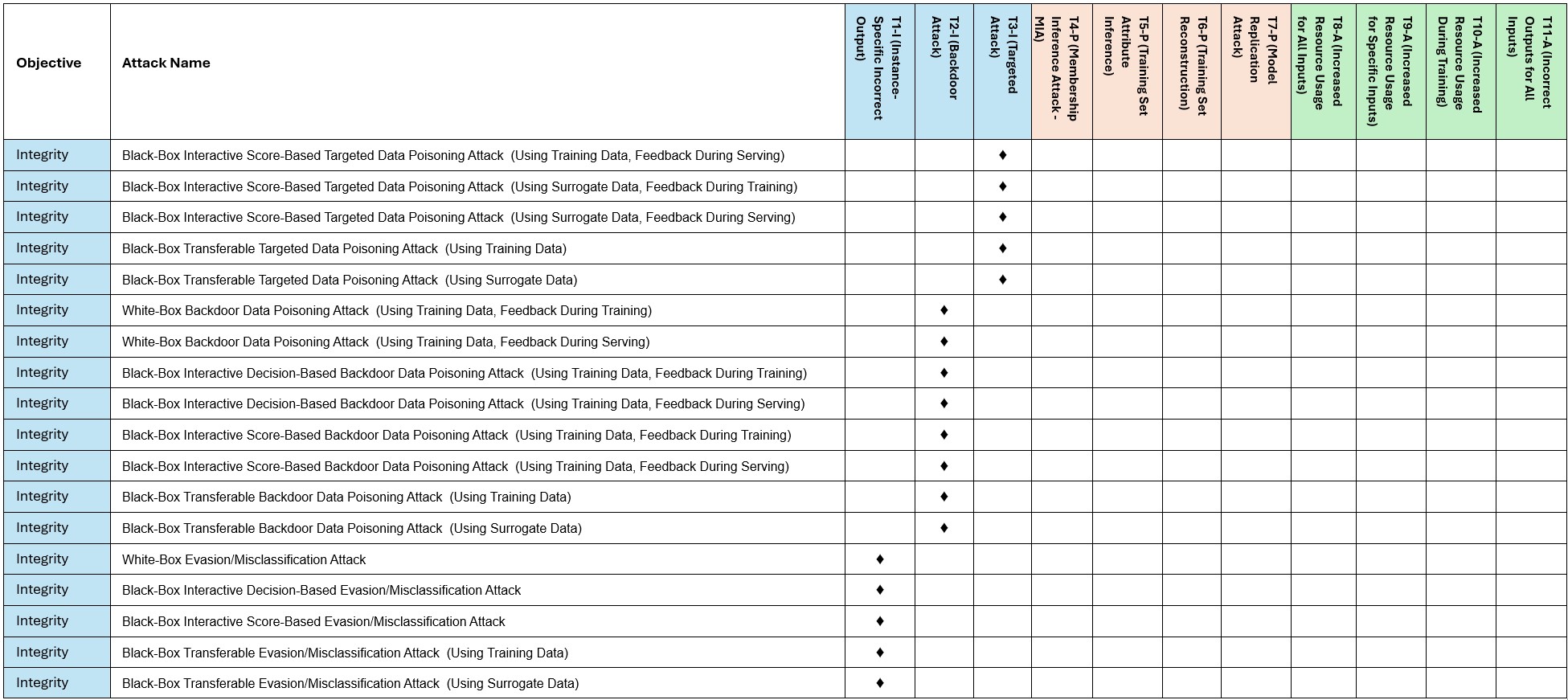}
        \caption{Attack impact mapping (3/3).}
        \label{AImp3}
    \end{figure*}

\end{landscape}

\clearpage

\onecolumn
\section{Attacker Feasibility Factors}
\label{Capabilities List}
\noindent The following table contains the feasibility factors possibly obtained by an attacker to execute different AML attacks and the relevant question from the system profiling questionnaire.
\vspace{10pt}

\renewcommand{\arraystretch}{1.2} 
\setlength{\tabcolsep}{5pt} 
\begin{longtable}{|c|p{4cm}|p{10cm}|}
    \hline
    \textbf{\#} & \textbf{Feasibility} & \textbf{Question} \\
    \hline
    1  & \textbf{Manipulating the Training Data} & How easy is it for the threat actor to compromise the data used for training or retraining the model? \\
    \hline
    2  & \textbf{Label Manipulation (Label Access)} & How easy is it for the threat actor to compromise the labeling procedure (i.e., causing incorrect labels to be included in the training set)? \\
    \hline
    3  & \textbf{Feature Engineering} & Do you perform any type of manual feature engineering? \\
    \hline
    4  & \textbf{Performing Training Data Validation} & What level of training data validation (e.g., cleaning, filtering, OOD/anomaly detection) is performed? \\
    \hline
    5  & \textbf{Performing Testing Data Validation} & What level of data validation (e.g., cleaning, filtering, OOD/anomaly detection) is performed at inference time (when the system is already deployed)? \\
    \hline
    6  & \textbf{Manipulating the Model's Parameters} & How secure are your model’s parameters against unauthorized changes? (e.g., sending malicious updates) \\
    \hline
    7  & \textbf{Model Retraining Frequency} & How often do you retrain the model? \\
    \hline
    8  & \textbf{Model Feedback (Training)} & What feedback is accessible to the threat actor during the model's training process? \\
    \hline
    9  & \textbf{Model Feedback (Serving)} & What is the feedback provided to model users at serving time? \\
    \hline
    10 & \textbf{End2End/Transfer Learning} & When retraining the model, do you retrain it from scratch or perform fine-tuning or transfer learning? \\
    \hline
    11 & \textbf{Model Online Evaluation} & Do you conduct online evaluation of your models? Does this include A/B testing? \\
    \hline
    12 & \textbf{Trigger the Attack Digital} & How easy is it for the threat actor to manipulate the model’s inputs digitally at serving time? (e.g., via an accessible API) \\
    \hline
    13 & \textbf{Manipulating the Environment (Trigger the Attack Physical)} & How easy is it for the threat actor to manipulate the model’s inputs physically at serving time? (e.g., placing an adversarial object in front of a sensor) \\
    \hline
    14 & \textbf{Access to Systems' Metrics - Training} & How easy is it for the threat actor to access systems' metrics during training time? (e.g., time, energy consumption, CPU) \\
    \hline
    15 & \textbf{Access to Systems' Metrics - Serving} & How easy is it for the threat actor to access systems' metrics at serving time? (e.g., time, energy consumption, CPU) \\
    \hline
    16 & \textbf{Training Data Knowledge} & How easy is it for the threat actor to obtain training data samples used for training the model? \\
    \hline
    17 & \textbf{Surrogate Data Knowledge} & How easy is it for the threat actor to obtain training data samples used for training the model or reference data similar to the data used for training the model? \\
    \hline
    18 & \textbf{Model Data Knowledge} & How much information does the threat actor have about the model’s design, training process, or deployment setup? \\
    \hline

    \caption{Attacker Feasibility Factors}
\end{longtable}

\clearpage

\onecolumn
\section{Impacts}
\label{Impact List}
\noindent The following table presents various types of attacks categorized by their impact on system security, privacy, and availability. Each attack is described with its potential effects and an example scenario.
\vspace{10pt} 

\renewcommand{\arraystretch}{1.3} 
\setlength{\tabcolsep}{5pt} 

\begin{longtable}{|c|c|p{10cm}|}
    \hline
    \textbf{ID} & \textbf{Category} & \textbf{Impact Description} \\
    \hline
    T1  & Integrity & \textbf{T1. Instance-Specific Incorrect Output (T1-I):}  
    The system produces incorrect outputs for specific inputs, leading to unexpected or unintended behavior that only affects the attacker's input.  
    \textit{Example:} A document verification system misclassifies a manipulated passport image as valid, granting unauthorized access to the attacker. \\
    \hline
    T2  & Integrity & \textbf{T2. Backdoor Attack (T2-I):}  
    The system behaves as intended under normal conditions but produces predefined incorrect outputs when triggered by specific inputs.  
    \textit{Example:} An image classification model consistently classifies images with a hidden watermark as "authorized," enabling unauthorized access when the trigger is present. \\
    \hline
    T3  & Integrity & \textbf{T3. Targeted Attack (T3-I):}  
    The system's behavior is altered to produce incorrect outputs for a specific subset of inputs chosen by the attacker, impacting a targeted subset while leaving the rest unaffected.  
    \textit{Example:} An attacker manipulates a financial model so that transactions from a specific region are consistently flagged as high-risk, regardless of their actual attributes. \\
    \hline
    T4  & Privacy & \textbf{T4. Membership Inference Attack (T4-P):}  
    An attacker is able to determine whether a specific data point was included in the model’s training dataset.  
    \textit{Example:} An attacker identifies that a particular medical record was used to train a diagnostic model, compromising patient privacy. \\
    \hline
    T5  & Privacy & \textbf{T5. Training Set Attribute Inference (T5-P):}  
    An attacker learns about general patterns or characteristics of the training data, exposing sensitive or private dataset attributes.  
    \textit{Example:} An attacker infers that a training dataset primarily includes data from a specific demographic group. \\
    \hline
    T6  & Privacy & \textbf{T6. Training Set Reconstruction (T6-P):}  
    An attacker manages to reconstruct samples from the training dataset, exposing private or proprietary data.  
    \textit{Example:} An attacker reconstructs sensitive training data, such as handwritten digits or personal emails, from a language model. \\
    \hline
    T7  & Privacy & \textbf{T7. Model Replication Attack (T7-P):}  
    An attacker replicates the model’s functionality, exposing intellectual property or gaining a competitive advantage.  
    \textit{Example:} An attacker trains a replica of a proprietary image classification model with nearly identical behavior. \\
    \hline
    T8  & Availability & \textbf{T8. Increased Resource Usage for All Inputs (T8-A):}  
    The system consumes significantly more resources, such as energy or computational time, for multiple inputs during inference, increasing operational costs and reducing efficiency.  
    \textit{Example:} Every query to a model takes twice as long to process, slowing response times and increasing cloud costs. \\
    \hline
    T9  & Availability & \textbf{T9. Increased Resource Usage for Specific Inputs (T9-A):}  
    The system consumes excessive resources for specific inputs, causing localized inefficiencies without affecting other users.  
    \textit{Example:} A single input forces a text generation model to perform excessive computations, significantly delaying its response. \\
    \hline
    T10 & Availability & \textbf{T10. Increased Resource Usage During Training (T10-A):}  
    The system requires significantly more resources, such as energy or computational time, during training, increasing costs and delaying deployment.  
    \textit{Example:} The model takes twice as long to converge during training, increasing costs. \\
    \hline
    T11 & Availability & \textbf{T11. Incorrect Outputs for All Inputs (T11-I):}  
    The system produces incorrect outputs for multiple inputs, affecting many users and interactions.  
    \textit{Example:} A recommendation system misclassifies every transaction, assigning high-risk transactions to low-risk clients and vice versa. \\
    \hline
    \caption{Attack Impact List}
\end{longtable}

\clearpage

\section{Zeroing (Drop-Out) Rules for Infeasible Attacks}
\label{zeroingAppendix}
As part of the \textbf{scoring process}, the capabilities defined by the \textbf{use-case profiling} are taken into account. In some cases, a specific attack may initially receive a feasibility score but is ultimately \textbf{infeasible} due to the constraints of the system. For example, a \emph{white-box clean-label poisoning attack} cannot occur if the model is not a \emph{white-box} system.

To systematically handle such cases, we define a set of \textbf{zeroing rules} that explicitly \textbf{invalidate (drop-out) anomalous attacks} when certain feasibility conditions are not met.

\subsection{Feasibility-Based Constraints}
The following use case parameters are some of the factors ~\ref{Capabilities List} used to assess feasibility:

\paragraph{Model Feedback (Training)} 
\textit{What feedback is accessible to the threat actor during the model's training process?} \\
Feedback refers to any information the adversary can access regarding the model’s behavior during training.
\begin{itemize}
    \item \textbf{Full access to the model’s flow}
    \item \textbf{Score-based}
    \item \textbf{Decision-based}
    \item \textbf{No feedback}
\end{itemize}

\paragraph{Model Feedback (Serving)} 
\textit{What feedback is provided to users at serving time?}
\begin{itemize}
    \item \textbf{Full access to the model’s flow}
    \item \textbf{Score-based}
    \item \textbf{Decision-based}
    \item \textbf{No feedback}
\end{itemize}

\paragraph{Knowledge of Model Data} 
\textit{How much information does the threat actor have about the model’s design, training process, or deployment setup?}
\begin{itemize}
    \item \textbf{Complete knowledge}
    \item \textbf{Known architecture}
    \item \textbf{Hyperparameters only}
    \item \textbf{Algorithm only}
    \item \textbf{Task only}
    \item \textbf{Unknown}
\end{itemize}

\subsection{Defined Zeroing Rules}
The following conditions will cause an attack’s score \( S(a) \) to be set to \textbf{zero (dropped out)}:

\paragraph{Score-Based Attacks}
\begin{itemize}
    \item If an attack relies on \textbf{score-based feedback}, but the system does not provide \textbf{"score-based"} or \textbf{"full access"} feedback in either the \textbf{training} or \textbf{serving} phase, the attack is \textbf{invalidated}.
\end{itemize}

\paragraph{White-Box Attacks}
\begin{itemize}
    \item If an attack is classified as \textbf{"white-box"}, but the \textbf{model knowledge} parameter does not indicate \textbf{"complete knowledge"}, the attack is \textbf{invalidated}.
    \item However, if \textbf{either} the \textbf{training feedback} or \textbf{serving feedback} is \textbf{"full access to model’s flow"}, the system is still treated as a \textbf{white-box} use case, and the attack remains feasible.
\end{itemize}

\subsection{Examples of Zeroed (Dropped-Out) Attacks}
\paragraph{White-Box Evasion/Misclassification Attack}
\begin{itemize}
    \item \textbf{Use Case Response:}
    \begin{itemize}
        \item \textbf{Model Feedback (Training)}: No feedback
        \item \textbf{Model Feedback (Serving)}: No feedback
        \item \textbf{Knowledge of Model Data}: "Task"
    \end{itemize}
    \item \textbf{Reason for Drop-Out}: The attack assumes a \textbf{white-box setting}, but the system provides \textbf{neither detailed feedback nor full model knowledge}, making it infeasible.
\end{itemize}

\paragraph{Black-Box Interactive Score-Based Targeted Model Poisoning Attack}
\begin{itemize}
    \item \textbf{Use Case Response:}
    \begin{itemize}
        \item \textbf{Model Feedback (Training)}: No feedback
        \item \textbf{Model Feedback (Serving)}: Decision-based
    \end{itemize}
    \item \textbf{Reason for Drop-Out}: The attack relies on \textbf{score-based feedback}, but the system does not expose score outputs in either the training or serving stages.
\end{itemize}

By implementing these \textbf{zeroing rules}, we ensure that inherently infeasible attack scenarios are automatically eliminated from the risk assessment process. This prevents the assignment of nonzero scores to attacks that cannot realistically be executed, improving the accuracy of the adversarial risk assessment.

\clearpage
\newpage
\section{Article Query List}
\label{Articles Query List}
The following queries were used to scrape publications on adversarial machine learning attacks in Google Scholar.

\footnotesize
\begin{multicols}{2}
\begin{itemize}[leftmargin=*, itemsep=1pt]

\item Adversarial ML attack

\item Adversarial ML attack Integrity Cyber
\item Adversarial ML attack Integrity NLP
\item Adversarial ML attack Integrity Computer Vision
\item Adversarial ML attack Integrity Network
\item Adversarial ML attack Integrity Speech
\item Adversarial ML attack Integrity Healthcare
\item Adversarial ML attack Integrity Finance

\item Black-Box Integrity attack Cyber
\item Black-Box Integrity attack NLP
\item Black-Box Integrity attack Computer Vision
\item Black-Box Integrity attack Network
\item Black-Box Integrity attack Speech
\item Black-Box Integrity attack Healthcare
\item Black-Box Integrity attack Finance

\item White-Box Integrity attack Cyber
\item White-Box Integrity attack NLP
\item White-Box Integrity attack Computer Vision
\item White-Box Integrity attack Network
\item White-Box Integrity attack Speech
\item White-Box Integrity attack Healthcare
\item White-Box Integrity attack Finance

\item Adversarial ML attack Availability Cyber
\item Adversarial ML attack Availability NLP
\item Adversarial ML attack Availability Computer Vision
\item Adversarial ML attack Availability Network
\item Adversarial ML attack Availability Speech
\item Adversarial ML attack Availability Healthcare
\item Adversarial ML attack Availability Finance

\item Black-Box Availability attack Cyber
\item Black-Box Availability attack NLP
\item Black-Box Availability attack Computer Vision
\item Black-Box Availability attack Network
\item Black-Box Availability attack Speech
\item Black-Box Availability attack Healthcare
\item Black-Box Availability attack Finance

\item White-Box Availability attack Cyber
\item White-Box Availability attack NLP
\item White-Box Availability attack Computer Vision
\item White-Box Availability attack Network
\item White-Box Availability attack Speech
\item White-Box Availability attack Healthcare
\item White-Box Availability attack Finance

\item Adversarial ML attack Privacy Cyber
\item Adversarial ML attack Privacy NLP
\item Adversarial ML attack Privacy Computer Vision
\item Adversarial ML attack Privacy Network
\item Adversarial ML attack Privacy Speech
\item Adversarial ML attack Privacy Healthcare
\item Adversarial ML attack Privacy Finance

\item Black-Box Privacy attack Cyber
\item Black-Box Privacy attack NLP
\item Black-Box Privacy attack Computer Vision
\item Black-Box Privacy attack Network
\item Black-Box Privacy attack Speech
\item Black-Box Privacy attack Healthcare
\item Black-Box Privacy attack Finance

\item White-Box Privacy attack Cyber
\item White-Box Privacy attack NLP
\item White-Box Privacy attack Computer Vision
\item White-Box Privacy attack Network
\item White-Box Privacy attack Speech
\item White-Box Privacy attack Healthcare
\item White-Box Privacy attack Finance

\item Adversarial attack in cybersecurity
\item Adversarial attack in NLP
\item Adversarial attack in Computer Vision
\item Adversarial attack in Network
\item Adversarial attack in Speech
\item Adversarial attack in Healthcare
\item Adversarial attack in Finance

\item Adversarial attack on deep learning models
\item Adversarial attack on decision trees
\item Adversarial attack on ensemble models
\item Adversarial attack on standard ML models
\item Adversarial attack on unsupervised learning models
\item Adversarial attack regression models

\item Adversarial ML attack Evasion Cyber
\item Adversarial ML attack Evasion NLP
\item Adversarial ML attack Evasion Computer Vision
\item Adversarial ML attack Evasion Network
\item Adversarial ML attack Evasion Speech
\item Adversarial ML attack Evasion Healthcare
\item Adversarial ML attack Evasion Finance

\item Black-Box Evasion attack Cyber
\item Black-Box Evasion attack NLP
\item Black-Box Evasion attack Computer Vision
\item Black-Box Evasion attack Network
\item Black-Box Evasion attack Speech
\item Black-Box Evasion attack Healthcare
\item Black-Box Evasion attack Finance

\item White-Box Evasion attack Cyber
\item White-Box Evasion attack NLP
\item White-Box Evasion attack Computer Vision
\item White-Box Evasion attack Network
\item White-Box Evasion attack Speech
\item White-Box Evasion attack Healthcare
\item White-Box Evasion attack Finance

\item Adversarial ML attack Misclassfication Cyber
\item Adversarial ML attack Misclassfication NLP
\item Adversarial ML attack Misclassfication Computer Vision
\item Adversarial ML attack Misclassfication Network
\item Adversarial ML attack Misclassfication Speech
\item Adversarial ML attack Misclassfication Healthcare
\item Adversarial ML attack Misclassfication Finance

\item Black-Box Misclassfication attack Cyber
\item Black-Box Misclassfication attack NLP
\item Black-Box Misclassfication attack Computer Vision
\item Black-Box Misclassfication attack Network
\item Black-Box Misclassfication attack Speech
\item Black-Box Misclassfication attack Healthcare
\item Black-Box Misclassfication attack Finance

\item White-Box Misclassfication attack Cyber
\item White-Box Misclassfication attack NLP
\item White-Box Misclassfication attack Computer Vision
\item White-Box Misclassfication attack Network
\item White-Box Misclassfication attack Speech
\item White-Box Misclassfication attack Healthcare
\item White-Box Misclassfication attack Finance

\item Adversarial ML attack Resources Latency Cyber
\item Adversarial ML attack Resources Latency NLP
\item Adversarial ML attack Resources Latency Computer Vision
\item Adversarial ML attack Resources Latency Network
\item Adversarial ML attack Resources Latency Speech
\item Adversarial ML attack Resources Latency Healthcare
\item Adversarial ML attack Resources Latency Finance

\item Black-Box Resources Latency attack Cyber
\item Black-Box Resources Latency attack NLP
\item Black-Box Resources Latency attack Computer Vision
\item Black-Box Resources Latency attack Network
\item Black-Box Resources Latency attack Speech
\item Black-Box Resources Latency attack Healthcare
\item Black-Box Resources Latency attack Finance

\item White-Box Resources Latency attack Cyber
\item White-Box Resources Latency attack NLP
\item White-Box Resources Latency attack Computer Vision
\item White-Box Resources Latency attack Network
\item White-Box Resources Latency attack Speech
\item White-Box Resources Latency attack Healthcare
\item White-Box Resources Latency attack Finance

\item Adversarial ML attack Poisoning Cyber
\item Adversarial ML attack Poisoning NLP
\item Adversarial ML attack Poisoning Computer Vision
\item Adversarial ML attack Poisoning Network
\item Adversarial ML attack Poisoning Speech
\item Adversarial ML attack Poisoning Healthcare
\item Adversarial ML attack Poisoning Finance

\item Black-Box Poisoning attack Cyber
\item Black-Box Poisoning attack NLP
\item Black-Box Poisoning attack Computer Vision
\item Black-Box Poisoning attack Network
\item Black-Box Poisoning attack Speech
\item Black-Box Poisoning attack Healthcare
\item Black-Box Poisoning attack Finance

\item White-Box Poisoning attack Cyber
\item White-Box Poisoning attack NLP
\item White-Box Poisoning attack Computer Vision
\item White-Box Poisoning attack Network
\item White-Box Poisoning attack Speech
\item White-Box Poisoning attack Healthcare
\item White-Box Poisoning attack Finance

\item Adversarial ML attack Targeted Poisoning Cyber
\item Adversarial ML attack Targeted Poisoning NLP
\item Adversarial ML attack Targeted Poisoning Computer Vision
\item Adversarial ML attack Targeted Poisoning Network
\item Adversarial ML attack Targeted Poisoning Speech
\item Adversarial ML attack Targeted Poisoning Healthcare
\item Adversarial ML attack Targeted Poisoning Finance

\item Adversarial ML attack Backdoor Poisoning Cyber
\item Adversarial ML attack Backdoor Poisoning NLP
\item Adversarial ML attack Backdoor Poisoning Computer Vision
\item Adversarial ML attack Backdoor Poisoning Network
\item Adversarial ML attack Backdoor Poisoning Speech
\item Adversarial ML attack Backdoor Poisoning Healthcare
\item Adversarial ML attack Backdoor Poisoning Finance

\item Adversarial ML attack Untargeted Poisoning Cyber
\item Adversarial ML attack Untargeted Poisoning NLP
\item Adversarial ML attack Untargeted Poisoning Computer Vision
\item Adversarial ML attack Untargeted Poisoning Network
\item Adversarial ML attack Untargeted Poisoning Speech
\item Adversarial ML attack Untargeted Poisoning Healthcare
\item Adversarial ML attack Untargeted Poisoning Finance

\item Adversarial ML attack Model Poisoning Cyber
\item Adversarial ML attack Model Poisoning NLP
\item Adversarial ML attack Model Poisoning Computer Vision
\item Adversarial ML attack Model Poisoning Network
\item Adversarial ML attack Model Poisoning Speech
\item Adversarial ML attack Model Poisoning Healthcare
\item Adversarial ML attack Model Poisoning Finance

\item Adversarial ML attack Clean Label Poisoning Cyber
\item Adversarial ML attack Clean Label Poisoning NLP
\item Adversarial ML attack Clean Label Poisoning Computer Vision
\item Adversarial ML attack Clean Label Poisoning Network
\item Adversarial ML attack Clean Label Poisoning Speech
\item Adversarial ML attack Clean Label Poisoning Healthcare
\item Adversarial ML attack Clean Label Poisoning Finance

\item Adversarial ML attack Data Reconstruction Cyber
\item Adversarial ML attack Data Reconstruction NLP
\item Adversarial ML attack Data Reconstruction Computer Vision
\item Adversarial ML attack Data Reconstruction Network
\item Adversarial ML attack Data Reconstruction Speech
\item Adversarial ML attack Data Reconstruction Healthcare
\item Adversarial ML attack Data Reconstruction Finance

\item Black-Box Data Reconstruction attack Cyber
\item Black-Box Data Reconstruction attack NLP
\item Black-Box Data Reconstruction attack Computer Vision
\item Black-Box Data Reconstruction attack Network
\item Black-Box Data Reconstruction attack Speech
\item Black-Box Data Reconstruction attack Healthcare
\item Black-Box Data Reconstruction attack Finance

\item White-Box Data Reconstruction attack Cyber
\item White-Box Data Reconstruction attack NLP
\item White-Box Data Reconstruction attack Computer Vision
\item White-Box Data Reconstruction attack Network
\item White-Box Data Reconstruction attack Speech
\item White-Box Data Reconstruction attack Healthcare
\item White-Box Data Reconstruction attack Finance

\item Adversarial ML attack Membership Inference Cyber
\item Adversarial ML attack Membership Inference NLP
\item Adversarial ML attack Membership Inference Computer Vision
\item Adversarial ML attack Membership Inference Network
\item Adversarial ML attack Membership Inference Speech
\item Adversarial ML attack Membership Inference Healthcare
\item Adversarial ML attack Membership Inference Finance

\item Black-Box Membership Inference attack Cyber
\item Black-Box Membership Inference attack NLP
\item Black-Box Membership Inference attack Computer Vision
\item Black-Box Membership Inference attack Network
\item Black-Box Membership Inference attack Speech
\item Black-Box Membership Inference attack Healthcare
\item Black-Box Membership Inference attack Finance

\item White-Box Membership Inference attack Cyber
\item White-Box Membership Inference attack NLP
\item White-Box Membership Inference attack Computer Vision
\item White-Box Membership Inference attack Network
\item White-Box Membership Inference attack Speech
\item White-Box Membership Inference attack Healthcare
\item White-Box Membership Inference attack Finance

\item Adversarial ML attack Property Inference Cyber
\item Adversarial ML attack Property Inference NLP
\item Adversarial ML attack Property Inference Computer Vision
\item Adversarial ML attack Property Inference Network
\item Adversarial ML attack Property Inference Speech
\item Adversarial ML attack Property Inference Healthcare
\item Adversarial ML attack Property Inference Finance

\item Black-Box Property Inference attack Cyber
\item Black-Box Property Inference attack NLP
\item Black-Box Property Inference attack Computer Vision
\item Black-Box Property Inference attack Network
\item Black-Box Property Inference attack Speech
\item Black-Box Property Inference attack Healthcare
\item Black-Box Property Inference attack Finance

\item White-Box Property Inference attack Cyber
\item White-Box Property Inference attack NLP
\item White-Box Property Inference attack Computer Vision
\item White-Box Property Inference attack Network
\item White-Box Property Inference attack Speech
\item White-Box Property Inference attack Healthcare
\item White-Box Property Inference attack Finance

\item Model Extraction attack Cyber
\item Model Extraction attack NLP
\item Model Extraction attack Computer Vision
\item Model Extraction attack Network
\item Model Extraction attack Speech
\item Model Extraction attack Healthcare
\item Model Extraction attack Finance

\end{itemize}
\end{multicols}
\normalsize

\vspace{10pt} 

\clearpage
\onecolumn

\newpage
\section{High-Impact Conferences and Journals Included in Our Dataset}
In this section we present the conferences and journals we focused on when looking for relevant publications in the dataset construction pipeline:
\label{ConAndJou}
\renewcommand{\arraystretch}{1.2}
\begin{longtable}{|p{4cm}|p{10cm}|}
\hline
\textbf{Category} & \textbf{Conferences} \\
\hline
Machine Learning & AAAI, ICDM, ICLR, ICML, IJCAI, KDD, NeurIPS, ECML PKDD, ECAI, WSDM, EMNLP, CIKM, ICDE, KDD, PAKDD \\
\hline
Security & ACM CCS, NDSS, IEEE S\&P, Usenix, ESORICS, AsiaCCS, ACSAC, TrustCom, DBSec, ACM SAC, Privacy Enhancing Technologies \\
\hline
Computer Vision & CVPR, ECCV, ICCV, BMVC, WACV \\
\hline
Other & Interspeech, Recsys, SIGIR, WWW, ICDE \\
\hline
\caption{Conferences}
\end{longtable}

\renewcommand{\arraystretch}{1.2}
\begin{longtable}{|p{4cm}|p{10cm}|}
\hline
\textbf{Domain} & \textbf{Journals} \\
\hline
\textbf{Machine Learning} & ACM Computing Surveys (CSUR), ACM Transactions on Intelligent Systems and Technology (TIST), ACM Transactions on Knowledge Discovery from Data, Advances in Neural Information Processing Systems, International Journal of Applied Machine Learning and Computational Intelligence \\
\hline
\textbf{Security} & Computers \& Security, IEEE Transactions on Dependable and Secure Computing, IEEE Transactions on Information Forensics and Security, International Journal of Information Security Science, Privacy Enhancing Technologies \\
\hline
\textbf{Computer Vision and AI Applications} & Engineering Applications of Artificial Intelligence, Expert Systems with Applications, Frontiers of Computer Science, Neurocomputing, Transactions on Latest Trends in Artificial Intelligence \\
\hline
\textbf{Networking and Communications} & IEEE Communications Magazine, IEEE Network, IEEE Wireless Communications, IEEE Wireless Communications Letters, Journal of Communications and Networks \\
\hline
\textbf{Healthcare and IoT} & American Journal of Industrial Medicine, IEEE Transactions on Intelligent Transportation Systems, Internet of Things \\
\hline
\textbf{General Computer Science} & IEEE Access, IEEE Internet Computing, IEEE Transactions on Knowledge and Data Engineering, IEEE Transactions on Software Engineering, Information Fusion, Journal of Computing and Information Technology \\
\hline
\caption{Journals}
\end{longtable}

\vspace{10pt} 

\newpage

\clearpage
\section{Use Case Analysis}
\label{useCasesAppendix}
In this section we present the other five use cases evaluated by our framework:

\subsection{(1) Use Case: Malware Classification for Email Security}
\textbf{System Description:}
The evaluated system is an ML malware classification model designed to enhance email security by detecting threats such as phishing, business email compromise (BEC) and spam. The system operates in a multi-cloud deployment and analyzes email metadata and attachments to identify malicious content. The classification results help organizations filter harmful emails and mitigate potential cybersecurity risks, ensuring a safer communication environment.

\textbf{Threat Actor Description:} 
An external adversary with no direct internal access, the selected threat actor is, however, capable of interacting with the classifier by sending emails and observing whether they are being filtered or delivered.

\textbf{Top Identified Attacks and Analysis:}
\begin{enumerate}
    \item \textbf{Black-Box Interactive Decision-Based Evasion/Misclassification Attack} \\
    \textbf{Score:} 7.48 \\
    \textbf{Objective:} Integrity \\
    \textbf{Explanation:} The attacker sends multiple variations of the same email and observes the system's feedback, such as whether an email is flagged as malicious or delivered to the recipient. By iteratively refining the email content or attachments based on the system’s response, the attacker crafts malicious emails that evade detection, undermining the integrity of the email classification system.

    \item \textbf{Black-Box Transferable Evasion/Misclassification Attack (Using Surrogate Data)} \\
    \textbf{Score:} 6.867 \\
    \textbf{Objective:} Integrity \\
    \textbf{Explanation:} Using a surrogate model trained offline with similar publicly available datasets, the attacker generates adversarial emails or attachments designed to bypass the malware classification system. This attack exploits the transferability of adversarial examples, compromising the integrity of the system by allowing harmful emails to be delivered.

    \item \textbf{Black-Box Transferable Evasion/Misclassification Attack (Using Training Data)} \\
    \textbf{Score:} 6.481 \\
    \textbf{Objective:} Integrity \\
    \textbf{Explanation:} The attacker might gain access to training data (for example either through publicly available sources or by exploiting the system's retraining process, which might unknowingly incorporate emails submitted by the attacker). Although harder to obtain than general surrogate data, using the training data enables the attacker to build a surrogate model that more accurately mirrors the behavior of the target classifier. This surrogate model is then employed to generate adversarial examples specifically crafted to exploit the system's weaknesses. By submitting these carefully crafted emails, the attacker bypasses detection, compromising the classifier's integrity and undermining its reliability.

    \item \textbf{Black-Box Interactive Decision-Based Data Reconstruction Attack} \\
    \textbf{Score:} 6.222 \\
    \textbf{Objective:} Privacy \\
    \textbf{Explanation:} The attacker iteratively submits crafted emails while analyzing system responses to reconstruct sensitive data samples (emails) that were part of the training set. 
    This process might lead to the unauthorized disclosure of proprietary or user data and could expose the company to privacy regulatory violations.
    \item \textbf{Black-Box Interactive Decision-Based Membership Inference Attack} \\
    \textbf{Score:} 4.474 \\
    \textbf{Objective:} Privacy \\
    \textbf{Explanation:} By crafting and submitting a series of queries, the attacker determines whether specific samples are part of the model’s training dataset, potentially exposing sensitive or proprietary information.
\end{enumerate}

\textbf{Evaluation Results: Feedback from System Owners and Experts}

\noindent\textbf{System Owners' Feedback:}
The system owners agreed with the framework's identification of the top threats as the most critical. They highlighted the severity of the two highest-ranked integrity attacks, noting their significant potential to allow malicious emails to bypass detection and compromise system reliability. Privacy-related attacks were also deemed relevant, as they could expose sensitive information or proprietary data. Availability attacks, which were not included in the top attack list, were identified as less critical in this context, as their disruptive nature often alerts system owners to potential issues, enabling rapid response.

\noindent\textbf{Experts' Feedback:}
The AML experts rated the \emph{overall framework accuracy} at \textbf{9/10}, indicating good alignment between the identified risks and real-world scenarios. For \emph{top threat relevance}, the experts \emph{strongly agreed} with the identified threats, affirming their criticality across integrity and privacy objectives. The average scores for \emph{attack-specific accuracy} and \emph{attack-specific relevance} were \textbf{8.8/10} and \textbf{8.6/10}, respectively, highlighting the framework's effectiveness in capturing the severity and practical significance of the top-ranked attacks.
Notably, the evasion/misclassification attacks received the highest scores for both accuracy and relevance, reflecting their significant potential to compromise the integrity of the system by allowing malicious emails to bypass detection. Conversely, while the privacy-related attacks were rated as critical due to their potential to expose proprietary or user-submitted data, they received slightly lower relevance scores on average, as experts noted that the iterative nature of such attacks may make them more noticeable in practice. The availability attack, though technically feasible, was rated as the least seriou, aligning with the system owners' feedback that its disruptive nature often signals system owners to potential issues, making it less stealthy and impactful.

\subsection{(2) Use Case: Traffic Steering in O-RAN}

\textbf{System Description:}  
The evaluated system is an ML-based traffic steering (TS) application in the Open Radio Access Network (O-RAN) architecture. This application automates the prioritization and distribution of network traffic to optimize efficiency and user experience. By integrating user equipment (UE)-centric strategies instead of traditional cell-centric approaches, the system dynamically adjusts network conditions based on real-time predictions. The TS process begins by identifying UE with anomalous quality of experience (QoE) and dynamically steering traffic to enhance network performance and individual user QoE.

\textbf{Threat Actor Description:}  
The threat actor in this use case consists of malicious UE or a group of UE, such as a botnet, that can send manipulated traffic patterns to the system. These devices interact with the TS system by generating and submitting QoE-related data, which the system uses as input for traffic steering decisions.

\textbf{Top Identified Attacks and Analysis:}  
\begin{enumerate}
    \item \textbf{Black-Box Transferable Backdoor Clean-Label Poisoning Attack (Using Surrogate Data)} \\
    \textbf{Score:} 7.0 \\
    \textbf{Objective:} Integrity \\
    \textbf{Explanation:}  The attacker uses available surrogate data to train a surrogate model that mimics the TS system’s behavior. Using this model, the attacker crafts benign-looking traffic samples embedded with a hidden backdoor trigger, which retain their original, legitimate labels. These samples are submitted as part of the regular traffic, so they are included in the system’s retraining dataset. During inference, when the backdoor trigger is activated (e.g., with a specific combination of QoE parameters or device characteristics), the model performs a predefined behavior. For instance, the attacker could force the system to prioritize their devices at the expense of others or disrupt the fairness of resource allocation, severely undermining the system's integrity and network efficiency.

    \item \textbf{Black-Box Transferable  Clean-Label Poisoning Attack (Using Surrogate Data)} \\
    \textbf{Score:} 6.414 \\
    \textbf{Objective:} Availability \\
    \textbf{Explanation:} The attacker trains a surrogate model using publicly available data to mimic the behavior of the TS system. Using this model, traffic samples are crafted to appear legitimate but are subtly designed to distort the model’s prioritization logic when included in retraining. For example, these samples might distort the model’s understanding of QoE-related parameters, leading to inefficient or imbalanced resource allocation. As these samples are incorporated in the retraining dataset, the TS system’s decisions become increasingly misaligned with actual network conditions. This results in degraded of quality service, such as increased latency, inefficient traffic distribution, or suboptimal utilization of network resources.

    \item \textbf{Black-Box Interactive Decision-Based Evasion/Misclassification Attack} \\
    \textbf{Score:} 5.542 \\
    \textbf{Objective:} Integrity \\
    \textbf{Explanation:} The attacker interacts with the system by submitting multiple variations of traffic patterns and observing the resulting decisions. Through iterative feedback, the attacker refines these patterns to mislead the QoE detection mechanisms. This leads to suboptimal or unfair traffic prioritization, compromising network efficiency and user experience.

    \item \textbf{Black-Box Transferable Evasion/Misclassification Attack (Using Surrogate Data)} \\
    \textbf{Score:} 5.385 \\
    \textbf{Objective:} Integrity \\
    \textbf{Explanation:} The attacker trains a surrogate model using similar available data to generate adversarial patterns that mislead the TS system’s detection mechanisms. These patterns are then transmitted through the infected UEs, allowing the attacker to disrupt the model’s ability to prioritize effectively. For example, the attacker may mislead the system into under-prioritizing legitimate users or overloading specific network resources.

    \item \textbf{Black-Box Transferable Resource-Latency Attack (Using Surrogate Data)} \\
    \textbf{Score:} 3.438 \\
    \textbf{Objective:} Availability \\
    \textbf{Explanation:} In this attack, the adversary leverages surrogate data to train a model that mimics the behavior of the target system. Using this surrogate model, the attacker generates carefully crafted traffic patterns designed to maximize computational overhead and resource usage during inference. These patterns, once transferred to the target system, exploit its processing limitations by forcing the model to allocate excessive resources to process the adversarial inputs. This results in increased inference latency, reduced throughput, and degraded system responsiveness. 
    This attack compromises the availability of the TS system, disrupting its ability to operate efficiently under real-time demands.
\end{enumerate}

\newpage

\textbf{Evaluation Results: Feedback from System Owners and Experts:}

\noindent\textbf{System Owners' Feedback:}
The system owners fully agreed with the framework’s ranking of the top threats, emphasizing their alignment with real-world vulnerabilities in the traffic steering system.

\noindent\textbf{Experts' Feedback:}
The experts rated the \emph{overall framework accuracy} and \emph{top threat relevance} with perfect scores of \textbf{10/10} and \emph{strongly agree}, underscoring the framework’s ability to identify and prioritize severe threats accurately. 
The framework was noted for covering all critical objectives (integrity, availability, and privacy), with no gaps identified.
For \emph{attack-specific accuracy and relevance}, the top-five attacks received consistently high scores, averaging \textbf{10/10} for both metrics, reflecting strong agreement with the ranking. 
Integrity and availability attacks were highlighted for their potential to manipulate traffic steering decisions, leading to unfair resource allocation and degraded user experience.

\subsection{(3) Use Case: Image Quality Ranking}
\textbf{System Description:}
The evaluated system is an image-ranking model designed to optimize product catalogs on an e-commerce platform. When multiple sellers upload images for the same product, the model evaluates each image and assigns it a quality rank. The image with the highest rank is selected to be the primary image for the product in the catalog, ensuring that buyers see the most representative and appealing image.
Quality ranking considers various aspects, including the visual clarity of the image and the absence of promotional content such as watermarks, URLs, or advertising banners. 

\textbf{Threat Actor Description:}
The selected threat actor is a seller on the e-commerce platform. Sellers list their products, including descriptions, prices, and images. For products sold by multiple sellers, the platform aggregates listings under a single product in the main catalog.

\textbf{Top Identified Attacks and Analysis:}
\begin{enumerate}
    \item \textbf{Black-Box Interactive Decision-Based Evasion/Misclassification Attack} \\
    \textbf{Score:} 7.48 \\
    \textbf{Objective:} Integrity \\
    \textbf{Explanation:} Sellers upload multiple images of a product and observe the feedback provided by the system (checking whether an image is selected as the primary product catalog image). This iterative process allows the sellers to refine their submissions by testing different variations of adversarial images. By exploiting this trial-and-error approach, sellers can manipulate the model into prioritizing low-quality or promotional images. This undermines the integrity of the platform and damages its credibility by showcasing unsuitable images in the catalog.

    \item \textbf{Black-Box Transferable Evasion/Misclassification Attack (Using Surrogate Data)} \\
    \textbf{Score:} 7.229 \\
    \textbf{Objective:} Integrity \\
    \textbf{Explanation:} The seller leverages a surrogate model trained offline on a surrogate dataset to mimic the behavior of the ranking model. By generating adversarial examples with this surrogate model, the seller creates images designed to receive high-quality rankings when uploaded to the platform. This attack exploits the transferability property of adversarial examples, tricking the system into prioritizing unsuitable or misleading images, thereby compromising the platform’s integrity.

    \item \textbf{Black-Box Transferable  Clean-Label Poisoning Attack (Using Surrogate Data)} \\
    \textbf{Score:} 4.694 \\
    \textbf{Objective:} Availability \\
    \textbf{Explanation:} 
    The seller uses a surrogate model trained on a surrogate dataset to create benign-looking images that are designed to negatively impact the system when included in retraining data. These images are uploaded as part of product listings and, without determining their labels, may be included in future retraining datasets. The poisoned data misleads the model into favoring substandard images, disrupting the ranking system and diminishing user trust and satisfaction. 
    
    \item \textbf{Black-Box Interactive Decision-Based Model Extraction Attack (Using Surrogate Data)} \\
    \textbf{Score:} 2.682 \\ 
    \textbf{Objective:} Privacy \\ 
    \textbf{Explanation:} By leveraging surrogate data (such as existing images chosen to be presented in the catalog) to guide queries effectively, the seller iteratively interacts with the ranking system and observes its decisions to reconstruct a model that replicates the behavior of the platform's ranking model. This reverse-engineered model might reveal the underlying ranking criteria, allowing the seller to craft images that achieve higher rankings while compromising the confidentiality of the company’s decision-making processes.
    
    \item \textbf{Black-Box Interactive Decision-Based Membership Inference Attack} \\
    \textbf{Score:} 1.51 \\
    \textbf{Objective:} Privacy \\
    \textbf{Explanation:} The attacker leverages interactive feedback from the ranking system to determine whether specific images were part of its training data. By crafting and submitting carefully designed queries, the attacker observes and analyzes the system’s responses to infer the membership status of target images. This information could be used to uncover images relied upon by the system or to facilitate subsequent attacks targeting the system.
\end{enumerate}

\textbf{Evaluation Results: Feedback from System Owners and Experts:}

\noindent\textbf{System Owners' Feedback:}
The system owners agreed with the framework's identification of the top three threats as the most critical. They emphasized the severity of the two highest-ranked attacks, particularly highlighting their significant potential to undermine the system's functionality and platform integrity. 

\noindent\textbf{Experts' Feedback:}
The AML experts rated the \emph{total framework accuracy} with a score of \textbf{8/10}, reflecting its ability to effectively prioritize adversarial threats. For \emph{top threat relevance}, the experts expressed \emph{agreement}, affirming the criticality of the identified threats. \emph{Objective coverage} received strong assessments for \emph{integrity} and \emph{availability} (\emph{yes}), while \emph{privacy} was marked as \emph{partially} addressed. However, this partial coverage aligns with the system owners' feedback, where privacy was considered the least critical objective for this system. 
Additionally, experts evaluated the accuracy and relevance of the identified attacks individually. The \emph{attack-specific accuracy} and \emph{attack-specific relevance} metrics achieved average scores of \textbf{8.8/10} and \textbf{8.4/10} respectively, further validating the framework's ability to highlight threats that are both impactful and practical to address. These results collectively demonstrate the framework’s effectiveness in identifying and prioritizing adversarial risks while providing actionable insights for stakeholders.

\subsection{(4) Use Case: Beam Hopping in Satellite Communication}
\textbf{System Description:}  
A model that is deployed on a satellite to manage beam hopping for communication optimization. The model receives data on the current demand and communication load in each cell and decides which beams to activate, aiming to optimize network performance by reducing latency, minimizing interference, and conserving energy. 

\textbf{Threat Actor Description:}  
The chosen threat actor is a malicious UE or a group of such devices, similar to a botnet. The attacker can infect multiple UEs connected to the satellite but cannot control their locations. 

\textbf{Top Identified Attacks and Analysis:}
\begin{enumerate}
    \item \textbf{Black-Box Interactive Decision-Based Evasion/Misclassification Attack} \\
    \textbf{Score:} 7.389 \\
    \textbf{Objective:} Integrity \\
    \textbf{Explanation:} The attacker (having infected multiple devices (UEs), sends varying communication patterns from these devices to interact with the model. By observing which patterns result in suboptimal beam activations, the attacker iteratively  refines their inputs. This feedback loop enables the attacker to craft communication patterns that  mislead the model into activating inefficient beams or causing interference, ultimately degrading the network’s performance.

    \item \textbf{Black-Box Interactive Score-Based  Clean-Label Poisoning Attack (Using Surrogate Data, Feedback During Training)} \\
    \textbf{Score:} 6.41 \\
    \textbf{Objective:} Availability \\
    \textbf{Explanation:} The attacker uses surrogate data, such as publicly available traffic patterns or simulated datasets, to understand the data distribution and craft poisoned samples that appear legitimate and retain clean labels. These samples are introduced into the system’s retraining dataset via compromised UEs. By interacting with the system and analyzing feedback, such as how the model adapts its decisions during training, the attacker iteratively refines the poisoned samples. These poisoned samples influence the model’s learning process, leading to inefficient beam hopping decisions, degraded resource allocation, and reduced system availability in high-demand scenarios.

     \item \textbf{Black-Box Transferable Evasion/Misclassification Attack (Using Surrogate Data)} \\
    \textbf{Score:} 6.35 \\
    \textbf{Objective:} Integrity \\
    \textbf{Explanation:} The attacker collects publicly available or simulated data representing network traffic to train a surrogate model that approximates the beam hopping system's decision-making process. This data may include general network conditions, traffic patterns, or signal propagation characteristics. Using the surrogate model, the attacker crafts adversarial patterns, which are then transferred to the target system and transmitted through compromised UEs. These patterns lead to misclassification of inputs, such as mistaking low-priority traffic for high-priority traffic or selecting suboptimal beam configurations. The attack results in inefficient beam allocations, increased interference between beams, and degraded service quality for critical users.

    \item \textbf{Black-Box Resource-Latency Attack (based on systems' metrics)} \\
    \textbf{Score:} 6.127 \\
    \textbf{Objective:} Availability \\
    \textbf{Explanation:}  The attacker gains indirect access to systems' metrics, such as processing time or performance delays, by observing the system's response to crafted communication patterns (e.g., analyzing response delays, packet acknowledgments, or throughput variations). Based on this feedback, the attacker iteratively crafts adversarial traffic patterns designed to maximize resource usage. These patterns force the system to allocate excessive computational resources to beam hopping decisions, degrading its ability to process legitimate traffic in real time. The result is reduced network availability, particularly during peak demand, impacting service reliability for users.

    \item \textbf{Black-Box Transferable Resource-Latency Attack (Using Surrogate Data)} \\
    \textbf{Score:} 6.001 \\
    \textbf{Objective:} Availability \\
    \textbf{Explanation:} The attacker leverages publicly available datasets or simulated traffic patterns, such as historical network performance data, coverage maps, or signal propagation characteristics, to train a surrogate model that mimics the target beam hopping system's behavior. Using this surrogate model, the attacker crafts adversarial traffic patterns designed to exploit the system's resource allocation mechanisms. These patterns are then transmitted through compromised UEs, forcing the target system to allocate excessive computational resources to beam hopping operations. This results in increased latency and reduced availability for legitimate traffic, especially during high-demand periods, causing service degradation.

\end{enumerate}
\textbf{Evaluation Results: Feedback from System Owners and Experts :}

\noindent\textbf{System Owners' Feedback:}
The system owners agreed with the framework's prioritization of integrity and availability attacks as the most critical risks to the beam hopping system. They emphasized the severe consequences of these attacks in real time. Privacy attacks were deemed less significant for this system, aligning with their absence from the top risks.

\noindent\textbf{Experts' Feedback:}
The experts rated the \emph{overall framework accuracy} at \textbf{10/10} and expressed \emph{strong agreement} with the relevance of the top-ranked attacks, affirming their alignment with real-world scenarios.
All critical objectives—\emph{integrity}, \emph{availability}, and \emph{privacy}—were considered adequately addressed. For both \emph{attack-specific accuracy} and \emph{attack-specific relevance} the top-five attacks received very high scores, averaging \textbf{9.8/10}.
Integrity-focused attacks, such as the \emph{black-box interactive decision-based evasion/misclassification attack}, were highlighted as the most severe, with perfect \textbf{10/10} ratings across both metrics, reflecting their ability to directly degrade the system’s core functionality. Availability attacks, including \emph{resource-latency attacks}, also achieved high scores, underscoring their critical impact in a real-time system like beam hopping, where resource optimization is paramount. The expert’s feedback confirmed the framework’s ability to effectively identify and rank the most pressing threats, particularly in high-stake scenarios involving network efficiency and user satisfaction.

\subsection{(5) Use Case: Product Relevance Classification}  

\textbf{System Description:}  
An ML model assigns relevance scores to sellers' products on an e-commerce platform based on the textual descriptions provided by the sellers.

When a buyer searches for a product, the system filters and displays only the products that meet a specific relevance threshold. The displayed results are sorted by relevance score, ensuring that buyers view the most relevant items first. This relevance-based filtering enhances user experience by reducing irrelevant or misleading results.

\textbf{Threat Actor Description:}  
The chosen threat actor in this use case is a seller who describes their products.

\textbf{Top Identified Attacks and Analysis:}  
\begin{enumerate}
    \item \textbf{Black-Box Interactive Decision-Based Evasion/Misclassification Attack} \\
    \textbf{Score:} 5.984 \\
    \textbf{Objective:} Integrity \\
    \textbf{Explanation:} The seller iteratively submits variations of their product's descriptions, observing which descriptions achieve higher relevance scores (by searching products in the platform as a buyer). By refining their inputs, the attacker crafts product descriptions that manipulate the model into rank their products higher in the search results, bypassing the intended fairness and accuracy of the system.

    \item \textbf{Black-Box Transferable Evasion/Misclassification Attack (Using Surrogate Data)} \\
    \textbf{Score:} 5.304 \\
    \textbf{Objective:} Integrity \\
    \textbf{Explanation:} The attacker trains a surrogate model on similar available datasets (like existing product descriptions on the platforms). Using this model, the attacker generates adversarial descriptions intended to increase the relevance scores of their products, tricking the system into prioritizing their items while excluding genuinely relevant ones.

    \item \textbf{Black-Box Transferable  Clean-Label Poisoning Attack (Using Surrogate Data)} \\
    \textbf{Score:} 3.655 \\
    \textbf{Objective:} Availability \\
    \textbf{Explanation:} The attacker crafts benign-looking but malicious product descriptions using a surrogate model. These descriptions are included in the system's retraining data, distorting the model's prioritization logic over time. The resulting distortion in product rankings reduces the model's performance and the system ability to provide relevant results for buyers.

    \item \textbf{Black-Box Transferable Backdoor Clean-Label Poisoning Attack (Using Surrogate Data)} \\
    \textbf{Score:} 1.502 \\
    \textbf{Objective:} Integrity \\
    \textbf{Explanation:} The attacker uses a surrogate model to create backdoor triggers embedded in natural product descriptions. During inference, the trigger activates, causing the system to prioritize the attacker's products disproportionately. This manipulation undermines the system's integrity by distorting the rankings.

    \item \textbf{Black-Box Interactive Decision-Based Data Reconstruction Attack} \\
    \textbf{Score:} 1.468 \\
    \textbf{Objective:} Privacy \\
    \textbf{Explanation:} The attacker iteratively submits strategically crafted descriptions and monitors the system's responses (the positioning of the seller's product in the search results). By analyzing these responses, the attacker could reconstruct samples from the training data, potentially gaining competitive advantage or enabling further exploitation of the system.
\end{enumerate}

\textbf{Evaluation Results: Feedback from System Owners and Experts :}

\noindent\textbf{System Owners' Feedback:}
The system owners agreed with the framework’s identification of integrity-focused attacks as the most critical, aligning with their interest in maintaining fair competition and accurate product rankings on the platform.

\noindent\textbf{Experts' Feedback:}
The experts rated the \emph{overall framework Accuracy} as \textbf{8/10}, affirming that the ranking effectively captures real-world risks. For \emph{top threat relevance}, the experts provided a rating of \emph{agreement}. The experts noted that the framework effectively balanced the scoring across objectives, accurately reflecting the varying severity of risks specific to this use case.
On average, the \emph{attack-specific accuracy} and \emph{attack-specific relevance} metrics for the top-five attacks scored \textbf{9.4/10} and \textbf{8.2/10}, respectively. Integrity-focused attacks, such as the \emph{black-box interactive decision-based evasion attack}, achieved perfect scores of \textbf{10/10} for both metrics, underscoring their importance in maintaining the system's credibility and fairness.
The \emph{black-box interactive decision-based data reconstruction attack} received significantly lower relevance scores, averaging \textbf{2/10}, reflecting its limited practical impact in this context. However, this aligns with the framework’s assessment, which assigned this attack a very low score of \textbf{1.502}.

\end{document}